\documentclass[]{youtu} 

\usepackage{mathpazo}
\usepackage{graphicx}
\usepackage[numbers]{natbib}
\usepackage{url}            
\usepackage{booktabs}       
\usepackage{amsfonts}       
\usepackage{nicefrac}       
\usepackage{microtype}      
\usepackage{xcolor}         

\usepackage{xspace}
\usepackage{amssymb}
\usepackage{amsmath}
\usepackage{wrapfig}
\usepackage{algorithm}
\usepackage{algorithmic}

\usepackage{multicol}
\usepackage{multirow}
\usepackage{makecell}
\usepackage{tabularx}
\usepackage{adjustbox}
\usepackage{enumitem}
\usepackage[normalem]{ulem}

\usepackage{pifont}
\usepackage{color}
\usepackage{colortbl}

\usepackage[capitalize]{cleveref}
\crefname{section}{Sec.}{Secs.}
\Crefname{section}{Section}{Sections}
\Crefname{table}{Table}{Tables}
\crefname{table}{Tab.}{Tabs.}

\definecolor{lightgray}{rgb}{0.8, 0.8, 0.8}
\definecolor{lgray}{rgb}{0.66, 0.66, 0.66}

\definecolor{lblu_tab}{RGB}{225, 235, 246}
\definecolor{orange_vitad}{RGB}{222, 131, 68}
\definecolor{blue_vitad}{RGB}{106, 153, 208}
\definecolor{trajectory_green}{RGB}{126, 171, 85}
\definecolor{trajectory_yellow}{RGB}{245, 194, 66}
\definecolor{tab_others}{RGB}{235, 235, 235}
\definecolor{tab_ours}{RGB}{225, 235, 246}

\definecolor{whit_tab}{RGB}{255, 255, 255}
\definecolor{gray_tab}{RGB}{220, 220, 220}
\definecolor{oran_tab}{RGB}{252, 242, 237}
\definecolor{blue_tab}{RGB}{227, 240, 251}

\def \pzo {\phantom{0}} 
\newcommand{\cmark}{\ding{52}\xspace}%
\newcommand{\xmarkg}{\textcolor{lightgray}{\ding{56}}\xspace}%

\def\onedot{.\xspace}
\def\eg{\textit{e.g}\onedot}

\usepackage{listings}
\usepackage{etoolbox}
\makeatletter
\AfterEndEnvironment{algorithm}{\let\@algcomment\relax}
\AtEndEnvironment{algorithm}{\kern2pt\hrule\relax\vskip3pt\@algcomment}
\let\@algcomment\relax
\newcommand\algcomment[1]{\def\@algcomment{\footnotesize#1}}
\renewcommand\fs@ruled{\def\@fs@cfont{\bfseries}\let\@fs@capt\floatc@ruled
  \def\@fs@pre{\hrule height.8pt depth0pt \kern2pt}%
  \def\@fs@post{}%
  \def\@fs@mid{\kern2pt\hrule\kern2pt}%
  \let\@fs@iftopcapt\iftrue}
\makeatother

\def\method{Soul}
\def\dataset{Soul-1M}
\def\benchmark{Soul-Bench}

\title{{\method}: Breathe Life into Digital Human \\for High-fidelity Long-term Multimodal Animation}
\author{\quad
Jiangning Zhang\quad
Junwei Zhu\quad
Zhenye Gan\quad
Donghao Luo\quad
Chuming Lin\quad
FeiFan Xu\quad
Xu Peng\quad \\
Jianlong Hu\quad
Yuansen Liu\quad
Yijia Hong\quad
Weijian Cao\quad
Han Feng\quad
Xu Chen\quad
Chencan Fu\quad \\
Keke He\quad
Xiaobin Hu\quad
Chengjie Wang\quad 
}
\affiliation{
YoutuVideo-Soul Team \\
}

\date{December 15, 2025}
\projectt{https://zhangzjn.github.io/projects/Soul/}{Soul Website}
\modell{https://huggingface.co/APRIL-AIGC/Soul}{Soul Model | Eval Suite}
\dataa{https://huggingface.co/datasets/APRIL-AIGC/Soul-Bench}{Soul-Bench}
\videoo{https://www.youtube.com/watch?v=vUzJl4cRdF0}{Soul Video}
\leader{Junwei Zhu}
\correspondence{186368@zju.edu.cn,~junweizhu@tencent.com}

\begin{document}

\abstract{
We propose a multimodal-driven framework for high-fidelity long-term digital human animation termed {\method}, which generates semantically coherent videos from a single-frame portrait image, text prompts, and audio, achieving precise lip synchronization, vivid facial expressions, and robust identity preservation. We construct {\dataset}, containing 1 million finely annotated samples with a precise automated annotation pipeline (covering portrait, upper-body, full-body, and multi-person scenes) to mitigate data scarcity, and we carefully curate {\benchmark} for comprehensive and fair evaluation of audio-/text-guided animation methods. The model is built on the Wan2.2-5B backbone, integrating audio-injection layers and multiple training strategies together with threshold-aware codebook replacement to ensure long-term generation consistency. Meanwhile, step/CFG distillation and a lightweight VAE are used to optimize inference efficiency, achieving an 11.4$\times$ speedup with negligible quality loss. Extensive experiments show that {\method} significantly outperforms current leading open-source and commercial models on video quality, video-text alignment, identity preservation, and lip-synchronization accuracy, demonstrating broad applicability in real-world scenarios such as virtual anchors and film production. 

}

\maketitle


\vspace{-.1em}

\section{Introduction} \label{sec:introduction} 
Digital human animation has become pivotal in entertainment, education, and telepresence, demanding semantic consistency with text/audio, high-fidelity, high-resoltuion, and long-duration generation~\cite{survey,wans2v,kling}. However, existing methods struggle to balance these requirements, limiting real-world deployment. This work targets generating dynamic, multimodal-aligned human videos from multimodal inputs (single image + text + audio), bridging the gap between technical feasibility and practical utility.

\begin{figure}[tp]
    \centering
    \includegraphics[width=1.0\linewidth]{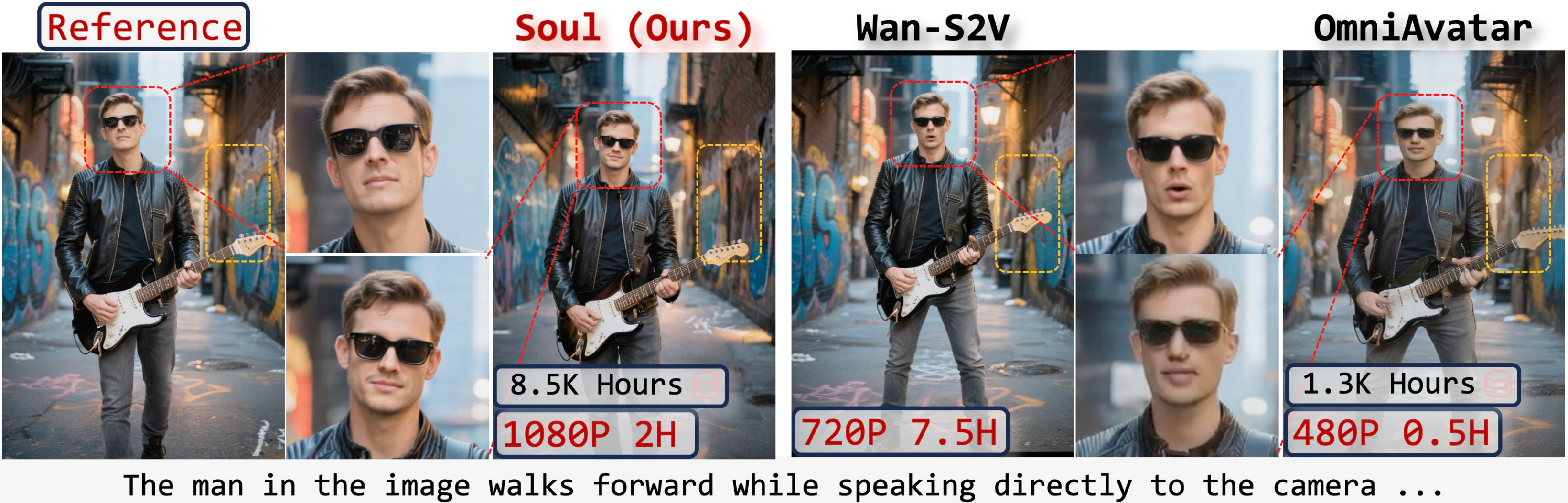}
    \caption{
    {\method} produces clearer results (\textcolor{red}{red} box), achieves stronger text consistency (walking shown in \textcolor{yellow}{yellow} box), and is faster (30s 1080P in 2-Hours) over recent Wan-S2V~\cite{wans2v} and OmniAvatar~\cite{omniavatar}. The data is derived from the AI-generated {\benchmark}, where the data is generated by the T2V model. 
    }
    \label{fig:teaser}
\end{figure}

Despite advances in video generation and audio-driven animation, three critical challenges remain unaddressed: 
\textbf{\textit{First}}, \textit{datasets lack comprehensive coverage and fine-grained annotations}: existing collections are biased toward single scenarios (\eg, only portraits) and lack detailed labels for actions, gestures, and scene dynamics, hindering model generalization. And the size of the training set is also important for practical performance. As shown in \cref{fig:teaser}, the latest Wan-S2V~\cite{wans2v} and OmniAvatar~\cite{omniavatar} (which only include 1.3K hours of training data) both fail to enable humans to "walk" effectively.
\textbf{\textit{Second}}, \textit{long-term inference suffers from identity drift and semantic degradation}, where latent feature shifts in extended generation break consistency, as models are trained on short clips but deployed for long sequences. It will also lead to a decline in the quality of the generated video, which may appear artifacts or blurriness. 
\textbf{\textit{Third}}, \textit{high-resolution (1080P) and high-fidelity generation conflicts with efficiency}: mainstream models either compromise quality for speed or require prohibitive computational resources, limiting real-time applications.

To tackle these challenges, we propose a holistic solution: 
\textbf{\textit{1)}} curate {\dataset} (see \cref{sec:dataset}) with diverse scenarios and fine-grained multimodal annotations to enhance model generalization; 
\textbf{\textit{2)}} design a long-term generation pipeline (see \cref{sec:long}) with pivotal frames, intra-clip overlap, and threshold-aware codebook replacement to mitigate identity drift and feature shift; 
\textbf{\textit{3)}} integrate audio-attention injection (see \cref{sec:model}) for precise lip-sync and text control; 
\textbf{\textit{4)}} optimize deployment for efficient high-resolution generation (see \cref{sec:infer}) to enhance its practical and commercial value. 
In summary, our contributions are threefold: 
\begin{itemize}
    \item To support high-fidelity, long-term multimodal animation, we propose a large-scale and diverse {\dataset} with fine-grained annotations by establishing a principled automated annotation pipeline.
    \item We introduce a novel multimodal {\method} framework with a discrete codebook, enabling long-term, identity-preserving, and high-fidelity human animation. By incorporating distillation and an efficient \textit{e}VAE, we significantly improve model efficiency with negligible performance degradation.
    \item We also construct a high-quality {\benchmark} to comprehensively and fairly evaluate audio-driven human animation methods. Results show that our approach not only produces high-quality 1080P generation with superior performance, but also achieves higher efficiency.
\end{itemize}

\section{Related Work} \label{sec:related_work}

\subsection{Video Generation Foundation Models}
The development of diffusion techniques has substantially improved image generation quality, exemplified by the SD~\cite{ldm,sdxl,sd3} and FLUX~\cite{flux,flux1kontext} series. This progress was soon extended to video generation, giving rise to works such as AnimateDiff~\cite{animatediff} and SVD~\cite{svd}, which advance video synthesis by expanding temporal modules. Transformer-based DiT architectures have become a standard component in subsequent video generation models; notably, Sora demonstrated a breakthrough in photorealistic generation, after which both open-source~\cite{goku, ltxvideo, stepvideot2v, skyreels, hunyuanvideo, wan, ultravideo, ultragen} and closed-source~\cite{vidu, ray2, runwaygen2, pixversev5, seedance1, waver, klingai, magi, veo3, sora2} models continued to improve generation quality. Wan2.1/2.2~\cite{wan} and HunyuanVideo~\cite{hunyuanvideo} have attracted particular attention for their strong results and well-maintained open ecosystems, and have been applied across many downstream tasks. Other work has focused on high-resolution and long-term video generation by introducing autoregressive modeling or improved attention mechanisms. Considering both performance and efficiency, we adopt Wan2.2-5B~\cite{wan} as our base video model.

\subsection{Audio-Driven Talking Human Generation}
Early audio-driven animation methods, such as SadTalker~\cite{sadtalker}, mainly targeted facial motion, typically using 3D Morphable Models (3DMM)~\cite{3dmm} to create animations and a GAN~\cite{gan}-based training pipeline, but they suffered from limited lip-sync accuracy and lackluster expression vividness. More recently, diffusion-based approaches have become dominant, and the animated subject has moved beyond the face to full-body and even multi-person scenarios, which offer greater practical value. Some U-Net-based methods~\cite{aniportrait,hallo,hallo2,sonic,synchrorama,magictalk} exploit pretrained denoising models to improve temporal consistency, yet they remain limited in identity preservation, audio conditioning control, and video extrapolation~\cite{survey}. Hallo3~\cite{hallo3} was the first to leverage DiT~\cite{dit}’s strong spatiotemporal modeling to handle more complex generation scenarios, a choice later adopted by AlignHuman~\cite{alignhuman}, Lookahead Anchoring~\cite{lookaheadanchoring}, and others. Subsequent works, \eg, FantasyID~\cite{fantasyid}, FantasyTalking~\cite{fantasytalking}, HunyuanVideo-Avatar~\cite{hunyuanvideoavatar}, InfinityHuman~\cite{infinityhuman}, build on powerful video foundation models such as CogVideoX~\cite{cogvideox}, Wan~\cite{wan}, HunyuanVideo~\cite{hunyuanvideo}, and Goku~\cite{goku}; by incorporating identity-reference networks or emotion control, these methods have achieved notable gains in consistency and expressiveness~\cite{omniavatar,echomimicv3,stableavatar,hunyuancustom,wans2v,omnihuman}. Other works~\cite{vexpress,echomimic,echomimicv2,midas,vividanimator} introduce additional pose sequences or keypoint inputs to enable finer and more vivid control of the body, at the cost of reduced applicability breadth. There are also promising results from approaches using 3DGS~\cite{egstalker}, Mixture-of-Experts (MoE)~\cite{muex}, and autoregressive (AR) paradigms~\cite{avatarsync}. Beyond single-person animation, several methods have been extended to multi-person or more general scenarios~\cite{fantasyportrait,multitalk,playmate2}. Despite these advances, existing work still faces substantial challenges in achieving fine-grained text/audio control for human-centric videos, robust audio-lip synchronization, large-range motion synthesis, and consistent identity preservation.

\subsection{Efficient Deployment}
For virtual-human generation applications, inference speed is critical for both user experience and operational cost. In the single-GPU end-to-end video inference setting, training-free attention implementations such as FlashAttention~\cite{flashattention1,flashattention2,flashattention3} and the SageAttention family~\cite{sageattention1,sageattention2,sageattention3} are widely adopted, while SpargeAttention~\cite{spargeattention} introduces sparse-attention mechanisms at the cost of higher hardware adaptation requirements. Some trainable methods address the performance drop caused by sparse attention and quantization through finetuning, \eg, VSA~\cite{vsa} and FPSAttention~\cite{fpsattention}. 
We investigate deployment from three angles: \textit{1)} seamless, training-free replacements for attention modules; \textit{2)} step and CFG distillation to fundamentally reduce the runtime cost of multi-step inference; and \textit{3)} a lightweight decoder design to achieve substantially lower end-to-end inference latency compared with mainstream approaches.

\section{{\method}: Multimodal Humam Animation} \label{sec:method}

\begin{figure*}[htp]
    \centering
    \includegraphics[width=1.0\linewidth]{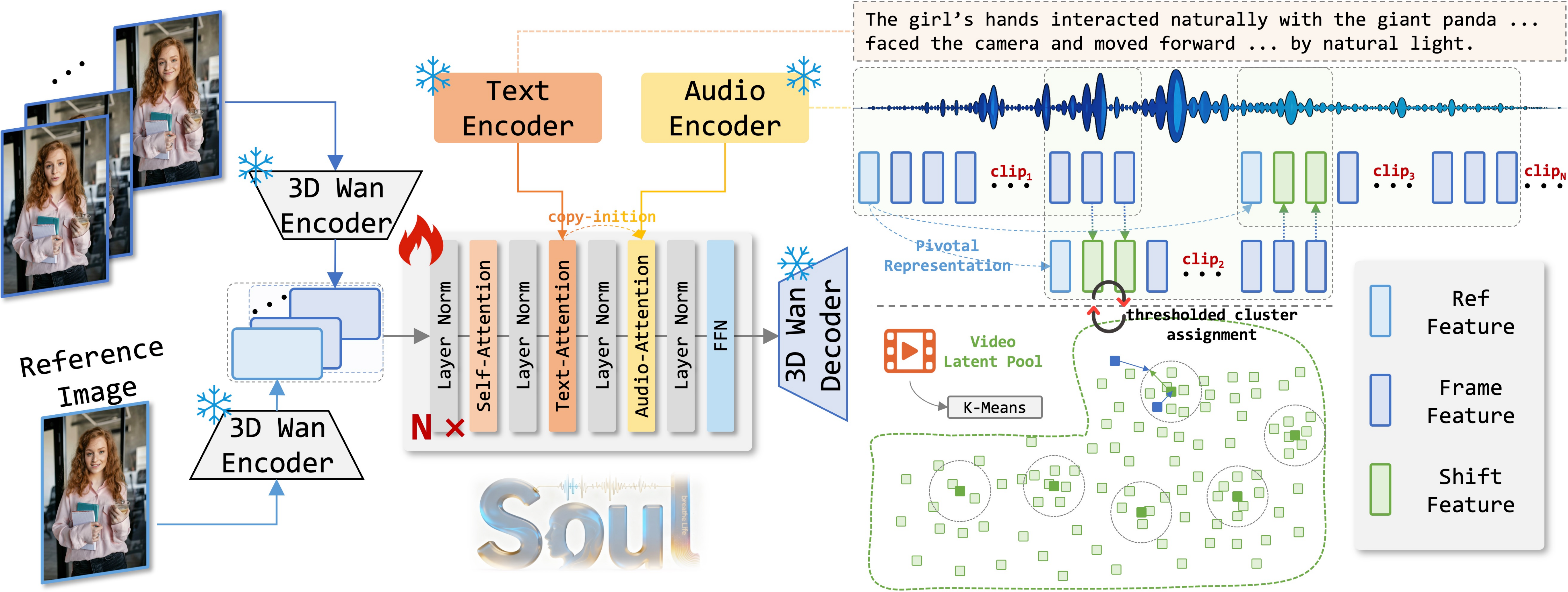}
    \caption{
    \textbf{Overview of {\method} for semantic-consistent and long-term multimodal-driven human video animation.} 
    }
    \label{fig:model}
\end{figure*}  

{\method} aims to generate dynamic animated videos whose actions and scenes are semantically consistent with text prompts and whose lip-sync and facial expressions match the audio-from a single-frame human image (which supplies the subject, background, and style) under multimodal guidance from text and audio, as illustrated in \cref{fig:model}. In particular, the audio modality supports various vocal types including general talking and singing, while the textual modality enables fine-grained control over the generated video by specifying task behaviors, gestures, scene details, shot scale, and camera motion. The system also supports prompt-driven specification of single- or multi-character scenarios for generalized inference. Moreover, we investigate long-term identity-preserving video generation and efficient deployment, and we construct a large-scale {\dataset} to support training of the proposed algorithms.

\subsection{Model Design} \label{sec:model}
\noindent\textbf{Video foundation model selection.} 
We adopt the recent Wan2.2-5B~\cite{wan} as our base video model. This model offers strong fidelity and robust general-scene generation capabilities, and its architecture is naturally advantageous for modeling human-world interactions. In addition, its larger downsampling factor enables direct generation of 1080P-resolution video, yielding substantially higher efficiency compared to open-source alternatives such as Wan2.2-1.3B~\cite{wan} and HunyuanVideo~\cite{hunyuanvideo}.

\noindent\textbf{Advancing DiT block for audio injection.} 
For the audio input, we pre-extract features using Whisper~\cite{whisper} and inject them by adding an Audio-Attention layer into the DiT blocks. Notably, this module is initialized from the original text-attention weights to accelerates model convergence.

\subsection{Long-term Human Video Generation} \label{sec:long}
\noindent\textbf{Pivotal representation.} 
By default we generate a single clip$_{n}$ containing 109 frames in pixel space, which corresponds to 28 frames in the latent space. As shown in \cref{fig:model}, we treat the first frame as a pivotal representation that is padded/duplicated to serve as the first frame of each clip; this pivotal frame provides consistent reference information for the clip-namely the subject identity, background, and style. During training we randomly sample one frame from the source video to serve as the reference frame to improve the model’s generalization to different human identities and backgrounds; this reference is used only as conditioning input and is not involved in the loss computation.

\noindent\textbf{Intra-clip overlap.} 
Consistency and plausibility between consecutive clips during long-term inference are crucial, and relying solely on a reference human image cannot ensure them. In particular, person motion driven by text or camera motion can make a clip’s ending diverge substantially from the pivotal representation. Therefore, when generating the next clip we copy the last few frames of the previous clip into the beginning of the current clip in the latent space (default: 2 frames). This overlap markedly improves semantic coherence across frames. During training, this strategy is applied with a probability of 50\%.

\noindent\textbf{Threshold-aware codebook replacement.} 
The two strategies above improve identity preservation and inter-frame consistency respectively, but we find that as inference length increases the generated video quality and scene coherence gradually degrade and can even become unusable (see \cref{fig:codebook}). We attribute this failure to a “latent feature shift”: during training the model only sees short-term (single) clips, whereas at inference it operates in a long-term (multiple-clip) regime. The core problem is that the preceding last few features used as conditioning are generated at inference time and therefore have a distributional mismatch with the real latent features of the first clip.

To mitigate this, we use all examples from {\dataset} (see \cref{sec:dataset}), pre-extract their latent features, and apply K-Means to form a set of clusters (we use 40K clusters by default) to construct a codebook that better matches the training distribution. We further introduce a thresholded cluster-assignment mechanism to replace each feature in the preceding frames via thresholded clipping. Concretely, for each feature we find its nearest cluster centroid: if the distance to that centroid is below a threshold, we keep the feature; if the distance exceeds the threshold, we move the feature toward the centroid and truncate its offset so that its distance to the centroid equals the threshold. This procedure keeps the preceding latent features close to the original training distribution while avoiding abrupt, wholesale replacements.

\noindent\textbf{Hybrid-modality training with negative guidance.} 
During training, we randomly mix in general videos without audio with a probability of 20\% to improve the model’s ability to maintain diverse scene types. These auxiliary samples include cartoons, animals, natural scenery and urban architecture; for them the audio channel is set to all zeros. In addition, we synthesize human-related failure cases using negative prompts and include these examples in training. We observe that this augmentation further enhances the model’s temporal consistency.

\subsection{Efficient Inference Deployment} \label{sec:infer}
To further improve user experience and reduce model cost, we optimize inference latency along two mutually compatible axes and ultimately achieve an 11.4-fold efficiency improvement compared with a naive inference implementation.

\noindent\textbf{Step and CFG Distillation.} 
Denoising steps have a large effect on output quality and by default we use 25 steps. Inspired by DMD2's distribution-matching distillation~\cite{DMD2}, we extend {\method} with simultaneous step and CFG (classifier-free guidance) distillation and remove the GAN loss. This modification substantially increases runtime speed while incurring only an acceptable accuracy degradation. As shown in \cref{tab:efficiency}, relative to the immediately preceding configuration, this yields a overall 7.5$\times$ improvement over the baseline model.

\begin{table}[tp]
    \caption{
    \textbf{Efficiency and performance of efficient \textit{e}VAE over official Wan2.2-5B-VAE.} Defalut 720$\times$1280 resolution on one GPU. 
    }
    \renewcommand{\arraystretch}{1.0}
    \setlength\tabcolsep{3.0pt}
    \resizebox{1.0\linewidth}{!}{
        \begin{tabular}{c|cc|cccc|ccc}
        \toprule[1.5pt]
        \multirow{2}{*}{VAE} & \multicolumn{2}{c}{Encoder} & \multicolumn{4}{c}{Decoder} & \multirow{2}{*}{PSNR$\uparrow$} & \multirow{2}{*}{SSIM$\uparrow$} & \multirow{2}{*}{LPIPS$\downarrow$} \\
        & Params. & MACs & Params. & MACs & Latency & Speedup & & & \\
        \hline
        Wan2.2-5B & 149.64M & 130.82T & 555.05M & 688.58T & 10.5796 & \pzo1.0$\times$ & 38.30 & 0.9567 & 0.0324 \\
        \textit{e}VAE-Wan2.2-5B-35M & 149.64M & 130.82T & \pzo34.97M & \pzo43.34T & \pzo1.3040 & \pzo8.1$\times$ & 37.14 & 0.9484 & 0.052\pzo \\
        \bottomrule[1.5pt]
        \end{tabular}
    }
    \label{tab:vae}
\end{table}

\noindent\textbf{Advancing Efficient VAE.} 
Beyond attention-centric optimizations, we observed that the decoder of Wan2.2-5B is relatively large: its parameter count and inference latency comprise a substantial fraction of the KD model's footprint. To address this, we designed a lightweight encoder/decoder variant, as summarized in \cref{tab:vae}. \textit{e}VAE-Wan2.2-5B-35M reduces decoder parameters and MACs from 550.05M/688.58T to 34.97M/43.34T while keeping reconstructed video quality within an acceptable range, \eg, LPIPS only increases from 0.0324 to 0.052.

\begin{figure*}[thp]
    \centering
    \includegraphics[width=1.0\linewidth]{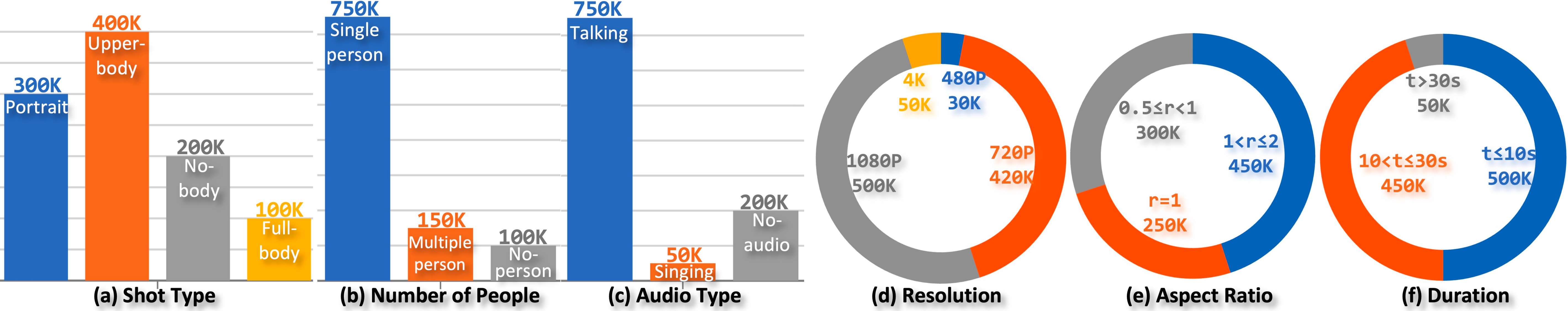}
    \caption{
    Statistical distributions of our \textbf{{\dataset}} from different perspectives.
    }
    \label{fig:analysis_train}
\end{figure*}  

\section{Curating {\dataset} Dataset} \label{sec:dataset}

\subsection{{\dataset} for Training}
\noindent\textbf{Human video data source.} 
To cover diverse application scenarios, we construct the large-scale {\dataset} from the following complementary perspectives: 
\textbf{\textit{1)}} Portraits: We leverage open-source datasets including VoxCeleb2~\cite{voxceleb2}, CelebV-Text~\cite{voxceleb2}, and VFHQ~\cite{vfhq}. These collections encompass celebrities, general-purpose footage, and interview-style recordings with varied duration distributions. Because faces occupy a large portion of the frame in these samples, they help strengthen the model’s ability to preserve lip motion and identity consistency. 
\textbf{\textit{2)}} Upper-body: To support use cases that focus on torso-level content, we gather indoor and interview-style videos that emphasize the upper body and hands, thereby improving consistency of generated upper-body poses and hand movements. 
\textbf{\textit{3)}} Full-body: To enable realistic full-body human animation, we collect large-scale full-body videos from general scenes covering diverse subjects, interactions, environments, lighting conditions, and atmospheres. These data target real-world, high-fidelity human animation. In addition, we augment this collection with manually annotated action data for specific gestures to enhance the expressiveness of intrinsic human motions under text control. 
\textbf{\textit{4)}} Multi-person scenes: We also include multi-person general-scene videos for mixed training to improve the model’s generalization and robustness in multi-person scenarios.

\noindent\textbf{Data filtering.}
Considering the variable quality of raw videos, we design an automated filtering pipeline consisting of: 
\textbf{\textit{1)}} Video resolution filtering: for high-definition target applications, we first remove clips below 480p at the source by retaining only videos whose short edge is greater than 480 pixels. 
\textbf{\textit{2)}} General-purpose model-based filtering: we detect scene cuts with PySceneDetect~\cite{pyscenedetect} in conjunction with DINOv2~\cite{dinov2}, remove clips that do not contain faces or have low face-detection confidence using RetinaFace~\cite{retinaface}, assess aesthetic quality with FineVQ~\cite{finevq}, delete clips containing heavy subtitles via PaddleOCR~\cite{paddleocr}, and finally run a multimodal large language model (MLLM)~\cite{qwen25vl} to comprehensively flag defects such as poor image quality, visible logos, or remaining subtitles. 
\textbf{\textit{3)}} Data-pool augmentation: we detect and temporally track each person’s keypoints with MMPose~\cite{mmpose}. For full-body videos, we generate upper-body crops and add them to the upper-body pool when the resulting crops still satisfy the resolution requirement. We also augment the pool with a subset of animation and pet videos. 
\textbf{\textit{4)}} Audio-video alignment: for each human video we track the largest face and apply SyncNet~\cite{syncnet} to evaluate audio-visual synchronization. Clips with persistent mismatch have their audio modality dropped while the video and its caption are retained for training. After this pipeline we obtain 300K portrait, 400K upper-body, and 100K full-body videos for training with totally 8.5K Hours.

\noindent\textbf{Automatic fine-grained labeling.}
\textbf{\textit{1)}} We adopt the open-sourced Qwen3-VL~\cite{qwen3,qwen25vl} as the base MLLM. We first determine human-centric event intervals based on “presence/absence of people” and “scene changes” (including scene changes caused by camera motion and transitions between stable scenes), and split each video into 4-5s sub-clips to ensure uniform annotation granularity and complete intervals.  
\textbf{\textit{2)}} For each sub-clip we design multi-type specialized prompts to comprehensively cover core attributes such as scene, shot type, human action, hand gesture, movement direction, camera movement, scene style, and lighting; multiple fine-grained templates are used to guide the model to output structured, detailed information.  
\textbf{\textit{3)}} In particular, for human action we decompose actions into body, head, face, hand, leg, etc., and design specialized prompts for these parts to ensure fine-grained action descriptions; we explicitly define object interaction to mean only physical-contact interactions, label any unconfirmable attributes as “none”, and employ both brief/caption-style prompts and detailed prompts to produce condensed and detailed outputs, balancing annotation efficiency and completeness.  
\textbf{\textit{4)}} Finally, we perform a secondary consistency check of the annotations using Qwen2.5-VL-72B~\cite{qwen25vl}, and remove examples where the video content contradicts the annotations (\eg, annotated “camera panning” but the shot is static, annotated “person walking” but the person shows no displacement), ensuring annotation accuracy.

\noindent\textbf{Statistical analysis from multiple perspectives.}
\cref{fig:analysis_train} explores the dataset across multiple dimensions. For \textbf{shot type} (a), \textit{Upper-body} (400K) is most frequent, followed by \textit{Portrait} (300K), \textit{No-body} (200K), and \textit{Full-body} (100K). In \textbf{number of people} (b), \textit{Single-person} (750K) dominates over \textit{Multiple-person} (150K) and \textit{No-person} (100K). For \textbf{audio type} (c), \textit{Talking} (750K) is prevalent, with \textit{Singing} (50K) and \textit{No-audio} (200K) trailing. \textbf{Resolution} (d) is led by 1080P (500K), then 720P (42W), 4K (50K), and 480P (30K). In \textbf{aspect ratio} (e), \(1 < r \leq 2\) (450K) is most common, followed by \(r = 1\) (250K) and \(0.5 \leq r < 1\) (300K). \textbf{Duration} (f) sees \(t \leq 10\,\text{s}\) (500K) and \(10 < t \leq 30\,\text{s}\) (450K) as major intervals, with \(t > 30\,\text{s}\) (50K) being the least. This multi-perspective overview reveals the dataset’s key characteristics.

\begin{figure}[t!]
    \centering
    \includegraphics[width=0.8\linewidth]{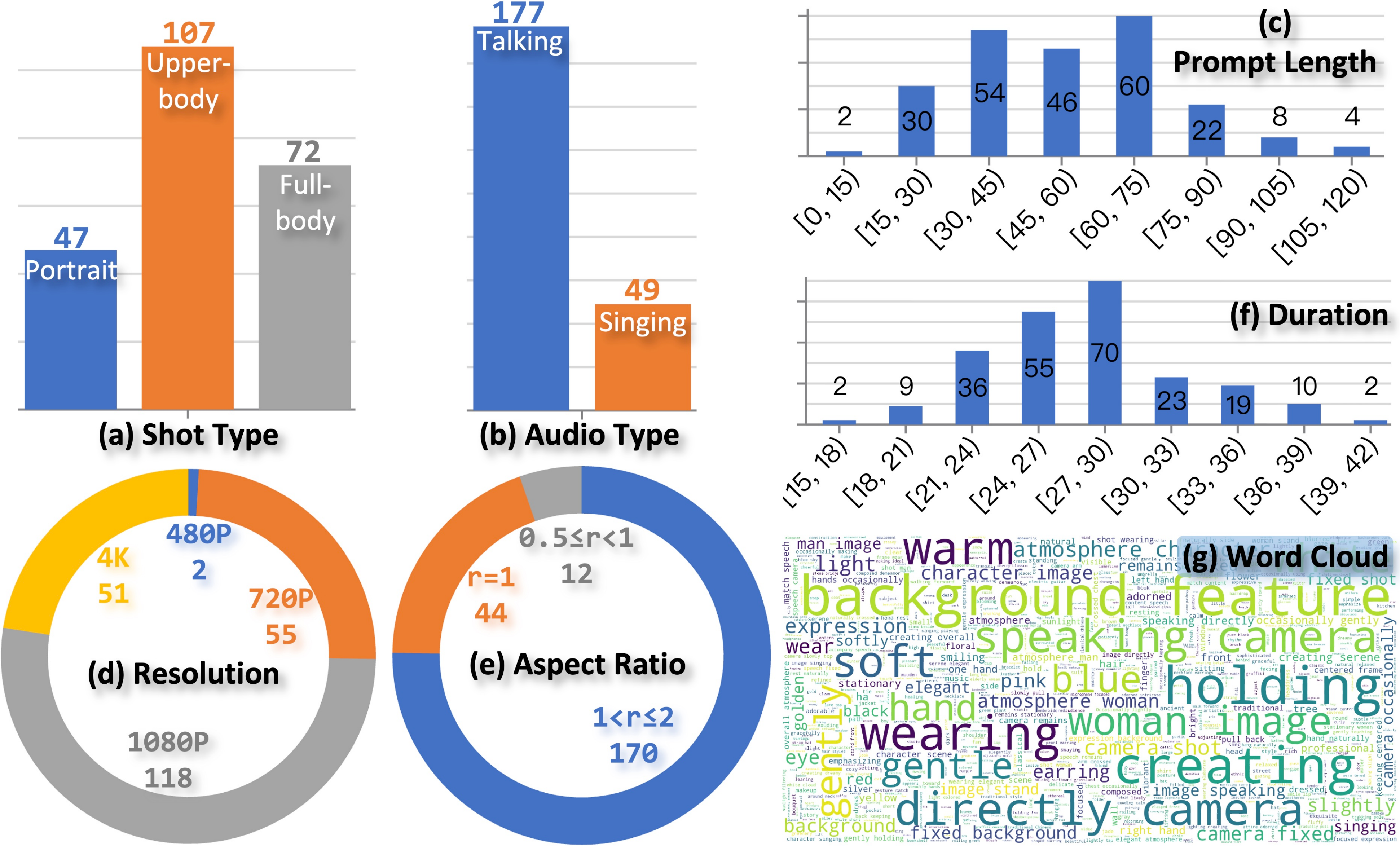}
    \caption{
    Statistical distributions of \textbf{{\benchmark}}. 
    }
    \label{fig:analysis_benchmark}
\end{figure}  

\begin{figure*}[t!]
    \centering
    \includegraphics[width=1.0\linewidth]{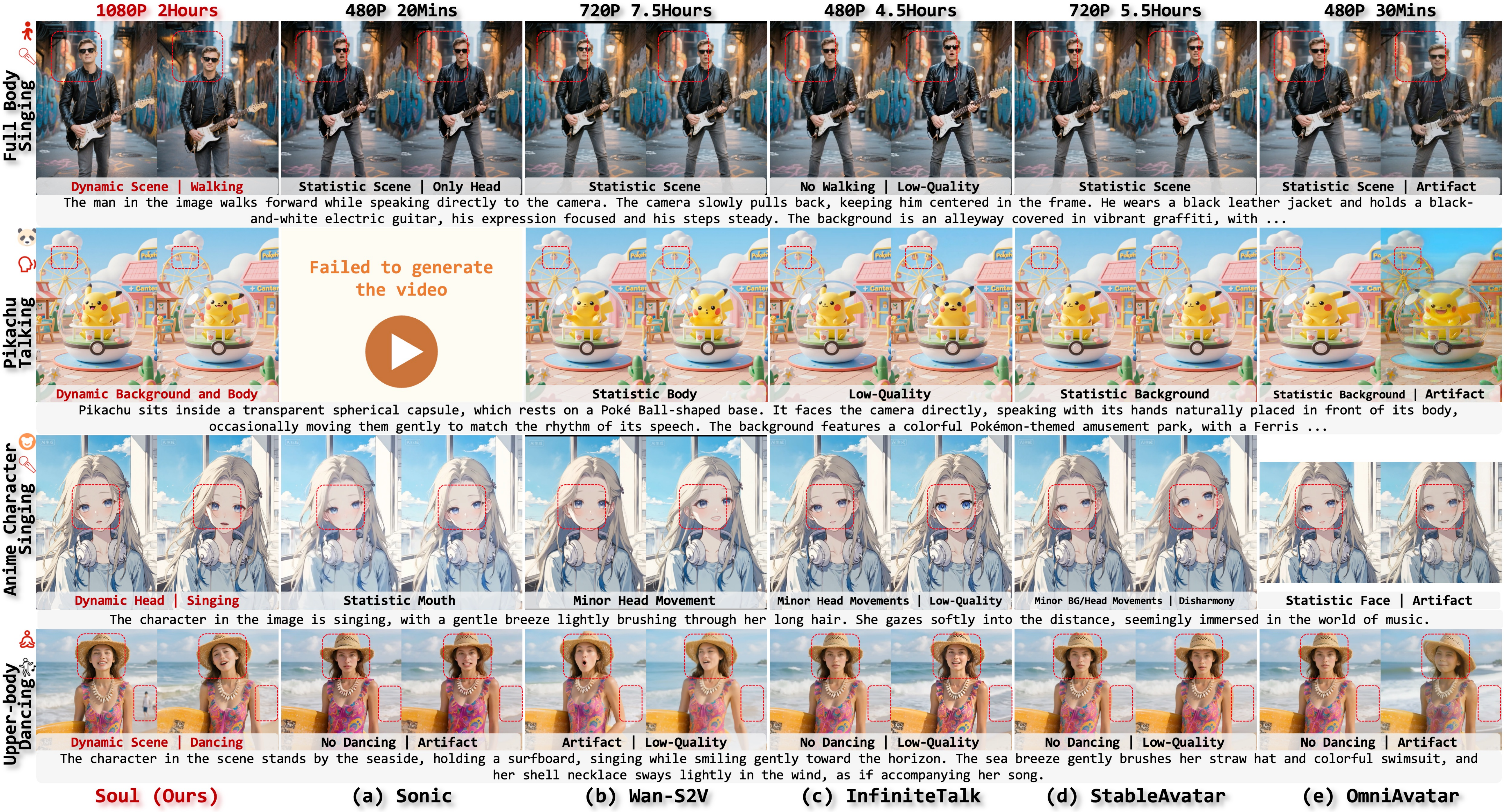}
    \caption{
    \textbf{Qualitative comparison with state-of-the-arts on {\benchmark}}. The timings are based on an average generation duration of 30s. Our {\method} can achieve strong semantic consistency and multi-scene generalization while preserving generation quality, with higher efficiency. The data is derived from the AI-generated {\benchmark}.
    }
    \label{fig:sota_qualitative}
\end{figure*}  

\begin{table*}[t!]
    \centering
    \caption{
    \textbf{Quantitative results with SoTAs on {\benchmark}.} \textbf{Bold} / \underline{underline} / \uwave{wavy line} for optimal / suboptimal / third-optimal metrics. For reference: LSE-C is 6.12 (training dataset) and Audio-Video Alignment is 23.19 (real videos)~\cite{av_align}. Values exceeding these thresholds lack significant distinguishability. Our {\method} comprehensively achieves significantly the best result.
    }
    \vspace{-1.0em}
    \label{tab:quantitative}
    \renewcommand{\arraystretch}{1.0}
    \setlength\tabcolsep{10.0pt}
    \resizebox{1.0\linewidth}{!}{
        \begin{tabular}{ccccccc}
        \toprule[0.1em]
        Method & Video-Text Consistence$\uparrow$ & LSE-D$\downarrow$ & LSE-C$\uparrow$ & Identity Consistence$\uparrow$ & Video Quality$\uparrow$ & Audio-Video Alignment$\uparrow$ \\
        \hline
        Sonic~\cite{sonic} & 4.57 & \underline{0.663} & \underline{7.80} & 0.613 & 68.58 & 0.191 \\
        Wan-S2V~\cite{wans2v} & 4.74 & 5.455 & 6.71 & \underline{0.750} & \uwave{71.22} & \textbf{0.330} \\
        InfiniteTalk~\cite{infinitetalk} & \uwave{4.75} & 2.313 & \textbf{8.48} & 0.609 & 68.53 & 0.211 \\
        StableAvatar~\cite{stableavatar} & \underline{4.77} & 3.948 & 4.05 & \uwave{0.733} & \underline{71.40} & \uwave{0.250} \\
        OmniAvatar~\cite{omniavatar} & \underline{4.77} & \uwave{1.009} & 5.84 & 0.497 & 67.24 & 0.225 \\
        \hline
        {\method} (Ours) & \textbf{4.85} & \textbf{0.130} & \uwave{6.82} & \textbf{0.763} & \textbf{72.60} & \underline{0.255} \\
        \toprule[0.1em]
        \end{tabular}
    }
\end{table*}

\subsection{{\benchmark}}
\begin{figure*}[t!]
    \centering
    \includegraphics[width=1.0\linewidth]{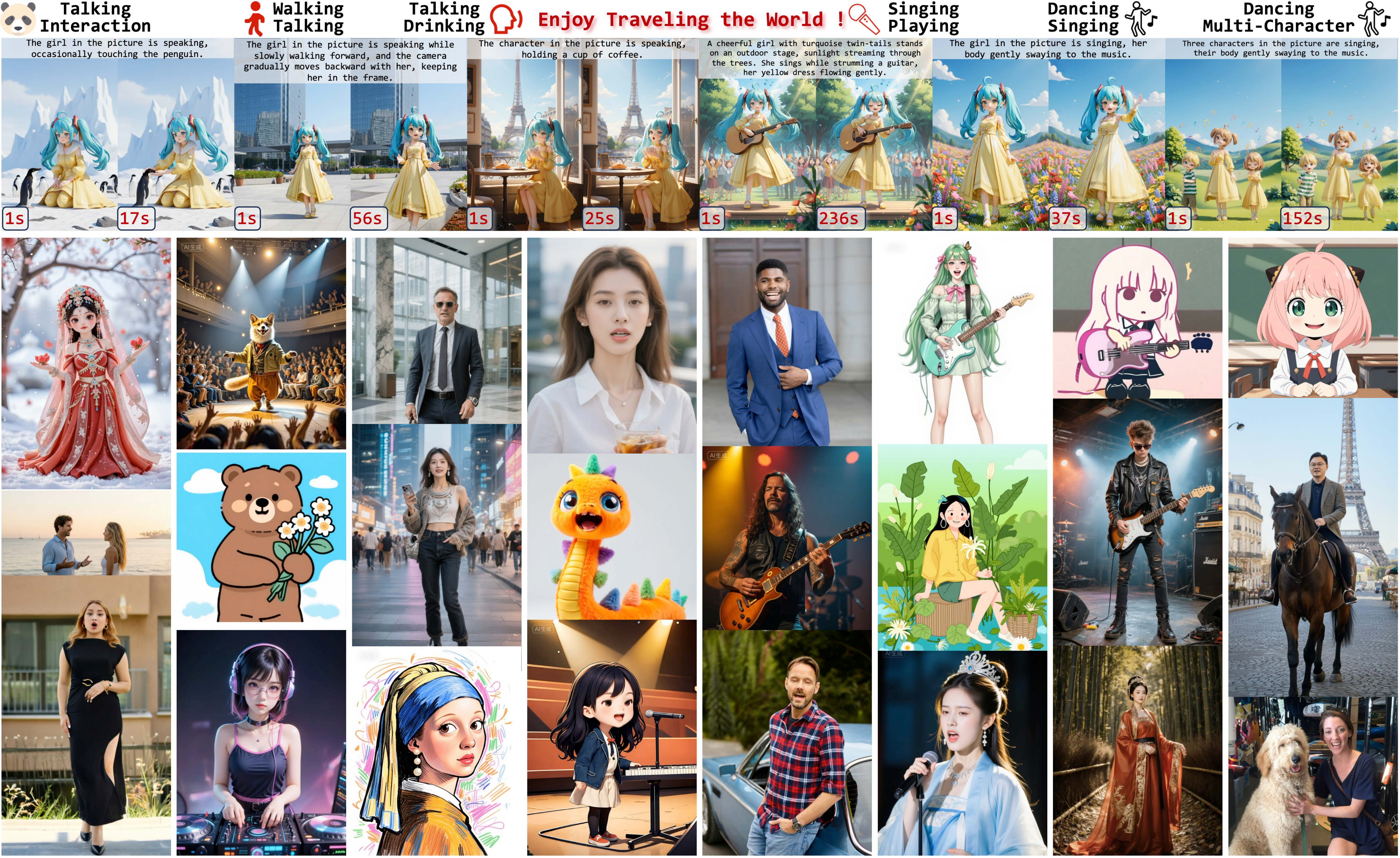}
    \caption{
    \textbf{Top:} Identity-consistent long-term animation across varying scenes with text and audio conditioning of our {\method}. \textbf{Bottom:} Diverse generative capabilities for practical applications of {\method} with samples from {\benchmark}. 
    }
    \label{fig:long}
\end{figure*}

\noindent\textbf{Manual dataset creation and selection.}
To comprehensively evaluate the performance and generalization of audio-guided human animation methods, we construct the {\benchmark} benchmark from practical application requirements using an automated-generation plus manual-verification pipeline. The benchmark contains 226 samples spanning portraits, upper-body, full-body, animation, and animal subjects, covering diverse durations, styles, genders, ages, accessories, environments, and interacting objects. The construction proceeds as follows: 
\textbf{\textit{1)}} Prompt and asset generation: we use an LLM~\cite{qwen3} to generate image prompts for HunyuanImage 3.0~\cite{hunyuanimage3}, produce talking text which is converted to audio by a TTS system~\cite{indextts2}, and generate scene text to guide the visual synthesis process. 
\textbf{\textit{2)}} Automatic plausibility filtering: a second LLM~\cite{deepseekr1} evaluates the realism and consistency of the synthesized assets. We iterate step 1 until the automated evaluator approves a target set size (500 entries in the automated stage). 
\textbf{\textit{3)}} Human verification: human annotators perform a further quality check on image fidelity, audio quality, and the reasonableness of the scene text, selecting only high-quality test samples. 
\textbf{\textit{4)}} Manual augmentation: we additionally curate and add real-world singing cases that are important in practice. 
After this pipeline we obtain 226 test samples (partially visualized in~\cref{fig:long}-Bottom) and the scene distribution statistics are shown in ~\cref{fig:analysis_benchmark}.

\noindent\textbf{Statistical analysis from multiple perspectives.}
\cref{fig:analysis_benchmark} covers multiple dimensions. For shot type (a), \textit{Upper-body} (107) is most common, followed by \textit{Full-body} (72) and \textit{Portrait} (47). Audio type (b) sees \textit{Talking} (177) dominate over \textit{Singing} (49). Prompt length (c) peaks at [60,75) (60). Resolution (d) is led by 1080P (118), then 720P (55), 4K (51), and 480P (2). Aspect ratio (e) has most cases in \(1 < r \leq 2\) (170), with \(r = 1\) (44) and \(0.5 \leq r < 1\) (12). Duration (f) is highest at [27,30) (70). This multi-angle overview reveals the dataset's key traits.

\noindent\textbf{Evaluation metrics for audio-guided human animation.}
We evaluate the generated videos using multiple foundation models from complementary perspectives: 
\textbf{\textit{1)}} ArcFace~\cite{arcface} to measure identity consistency between the source and synthesized subjects.  
\textbf{\textit{2)}} DINOv2~\cite{dinov2} to assess consistency in human-aware regions (\eg, body and face structural alignment).  
\textbf{\textit{3)}} FineVQ~\cite{finevq} to quantify perceptual/video quality of the outputs.  
\textbf{\textit{4)}} Qwen3-VL~\cite{qwen3,qwen25vl} to evaluate adherence to textual instructions (text-following capability).  
\textbf{\textit{5)}} SyncNet~\cite{syncnet} to evaluate audio-visual synchronization (lip-audio alignment).

\section{Experiments} \label{sec:exp}

\subsection{Experimental Setup}
\noindent\textbf{Baselines.} 
We compare against the latest open-source Sonic~\cite{sonic}, Wan-S2V~\cite{wans2v}, InfiniteTalk~\cite{infinitetalk}, StableAvatar~\cite{stableavatar}, and OmniAvatar~\cite{omniavatar}, as well as the closed-source HeyGen~\cite{heygen} and Kling-Avatar~\cite{kling}.

\noindent\textbf{Implementation detail.}
We train on {\dataset}, which contains 80W synchronized text-audio-video samples, with an additional 20W general videos without audio, covering portrait, half-body, and full-body shots. {\method} is built on Wan2.2-5B~\cite{wan}. A single inference by default generates a video clip of 109$\times$1088$\times$1920 resolution (all videos are scaled to an equivalent resolution of 1088$\times$1920 while maintaining the aspect ratio), and we employ the design in \cref{sec:long} for long-term video generation. We fully fine-tune the model on 64 GPUs using AdamW~\cite{adamw} with a learning rate of $2\times10^{-5}$. To improve training efficiency, we first train for 2 epochs at a resolution of 109$\times$720$\times$1280, and then fine-tune for 1 epoch at 1088$\times$1920 resolution.

\noindent\textbf{Video assessment.}
{\benchmark} uses the following metrics to evaluate different methods: 
\textbf{\textit{1)}} \textbf{Video-Text Consistence} evaluated by Qwen3-VL-235B-A22B-Instruct~\cite{qwen3,qwen25vl}; 
\textbf{\textit{2)}} \textbf{LSE-D} (Lip-Sync Error-Confidence) and \textbf{LSE-D} (Lip-Sync Error-Distance) to assess lip-audio synchronization using SyncNet-v2~\cite{latentsync}; 
\textbf{\textit{3)}} \textbf{Identity Consistence} by computing cosine similarity using ArcFace~\cite{arcface}; 
\textbf{\textit{4)}} \textbf{Video Quality} measured by the overall score of FineVQ~\cite{finevq}; 
\textbf{\textit{5)}} \textbf{Audio-Video Alignment} following the method in~\cite{av_align}. 
Additionally, we conduct a human study to evaluate overall video effectiveness: aesthetic visual quality, consistency with textual instructions, and lip-sync consistency.

\begin{table*}[t!]
    \caption{ 
    \textbf{Impact of different acceleration components on efficiency and performance.} 
    Defalut 129$\times$1088$\times$1920 resolution on one GPU. Speedup is relative to the Baseline of the first line. FA2: FlashAttention2; KD: Step and CFG Knowledgement Distillation; \textit{e}VAE: Our designed efficient \textit{e}VAE-Wan2.2-5B-35M.
    }
    \renewcommand{\arraystretch}{1.1}
    \setlength\tabcolsep{6.0pt}
    \resizebox{1.0\linewidth}{!}{
        \begin{tabular}{ccc|cc|cc|cc|cccccc}
        \toprule[1.5pt]
        \multirow{2}{*}{\makecell[c]{FA2}} & \multirow{2}{*}{\makecell[c]{KD}} & \multirow{2}{*}{\makecell[c]{\textit{e}VAE}} & \multicolumn{2}{c}{DiT} & \multicolumn{2}{c}{Decoder} & \multicolumn{2}{c}{Full Model} & \multirow{2}{*}{\makecell[c]{Video-Text \\ Consistence$\uparrow$}} & \multirow{2}{*}{\makecell[c]{LSE-D$\downarrow$}} & \multirow{2}{*}{\makecell[c]{LSE-C$\uparrow$}} & \multirow{2}{*}{\makecell[c]{Identity \\Consistence$\uparrow$}} & \multirow{2}{*}{\makecell[c]{Video \\ Quality$\uparrow$}} & \multirow{2}{*}{\makecell[c]{Audio-Video \\Alignment$\uparrow$}} \\
        & & & Latency & Speedup & Latency & Speedup & Latency & Speedup & & & & & & \\
        \hline
        \cmark & \xmarkg & \xmarkg & 960.75s & \pzo1.0$\times$ & 55.75s & 1.0$\times$ & 1019.2s & \pzo1.0$\times$ & 4.85 & 0.130 & 6.82 & 0.763 & 72.60 & 0.255 \\
        \hline
        \cmark & \cmark & \xmarkg & \pzo76.86s & 12.5$\times$ & 55.75s & 1.0$\times$ & \pzo135.3s & \pzo7.5$\times$ & 4.81 & 0.180 & 6.52 & 0.696 & 71.90 & 0.274 \\
        \cmark & \cmark & \cmark & \pzo76.86s & 12.5$\times$ & \pzo9.85s & 5.7$\times$ & \pzo\pzo89.4s & 11.4$\times$ & 4.83 & 0.144 & 6.52 & 0.702 & 71.68 & 0.272 \\
        \bottomrule[1.5pt]
        \end{tabular}
    }
    \label{tab:efficiency}
\end{table*}

\begin{figure}[t!]
    \centering
    \includegraphics[width=1.0\linewidth]{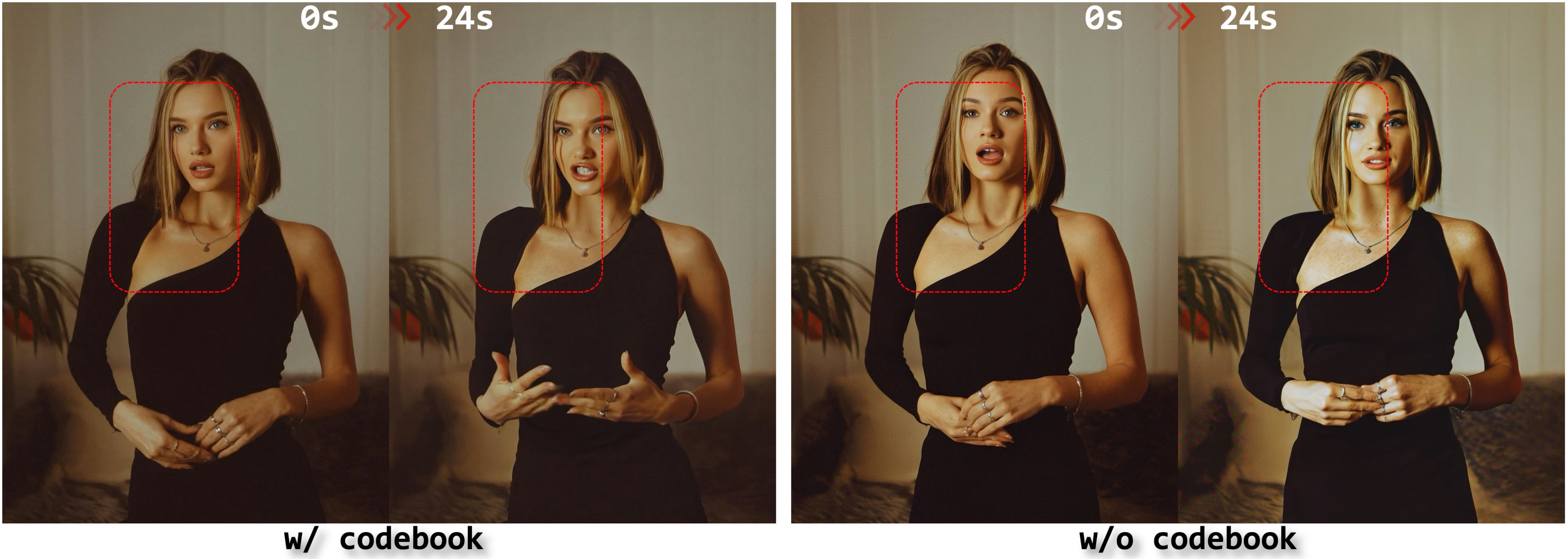}
    \caption{
    Over time, the approach without using the threshold-aware codebook is prone to color deviation and loss of details. The data is derived from the AI-generated {\benchmark}.
    }
    \label{fig:codebook}
\end{figure}

\subsection{Experimental Results}

\subsubsection{Comparison with Open-sourced SoTAs}
\cref{fig:sota_qualitative} shows qualitative comparisons with state-of-the-art (SoTA) methods. As can be seen, compared to competing approaches, our {\method} consistently generates semantically coherent, high-quality videos across single-person, environment-interaction, anime, and animal scenarios, while offering higher generation resolution and efficiency than the counterparts. \cref{tab:quantitative} reports the quantitative results: compared to competing methods, our {\method} achieves superior video-text alignment and audio-visual synchronization, and better preserves identity consistency.

\subsubsection{Ablation and Analysis.}
\noindent\textbf{Threshold-aware codebook replacement.} 
\cref{fig:codebook} intuitively illustrates that even in the presence of intra-clip overlap, the generated video may suffer unexpected degradation in visual quality as time progresses, and the threshold-aware codebook strategy introduced in \cref{sec:long} can effectively mitigate this issue. 

\noindent\textbf{Long-term generation.}
Thanks to the components in \cref{sec:long}, our method is better suited for long-term human video generation. \cref{fig:long}-Top shows that our method can generate videos of up four minutes in length: while preserving image quality, both subject identity and background consistency are well maintained, and smooth, seamless transitions across different scenes are achieved.

\noindent\textbf{Effect of KD and \textit{e}VAE.} 
\cref{tab:efficiency} demonstrates the effectiveness of KD and \textit{e}VAE-Wan2.2-5B-35M: compared to the non-deployed model, the overall system achieves an 11.4$\times$ speedup with negligible performance degradation.

\begin{table}[tp]
    \centering
    \caption{Human study with commercial products. \ding{192} Overall Naturalness, \ding{193} ID Consistency, \ding{194} Text Consistency, and \ding{195} Audio-Visual Synchronization.}
    \label{tab:human}
    \renewcommand{\arraystretch}{1.0}
    \setlength\tabcolsep{12.0pt}
    \resizebox{0.6\linewidth}{!}{
        \begin{tabular}{ccccc}
        \toprule[0.1em]
        Method & \ding{192} & \ding{193} & \ding{194} & \ding{195} \\
        \hline
        HeyGen~\cite{heygen} & 4.07 & 3.54 & 3.82 & \textbf{4.20} \\
        Kling-Avatar~\cite{kling} & 3.93 & 3.86 & 3.90 & 4.05 \\
        \hline
        Ours & \textbf{4.17} & \textbf{4.00} & \textbf{4.11} & \textbf{4.20} \\
        \toprule[0.1em]
        \end{tabular}
    }
\end{table}

\noindent\textbf{Comparison with commercial products.}
Further, we conducted a human evaluation via API calls comparing our method to the commercial Kling-Avatar~\cite{kling} and HeyGen~\cite{heygen} in \cref{tab:human}. Specifically, using the same inputs we generated 100 videos per model (as of October 31) and hired 10 professional video raters to score them on four dimensions: \ding{192} overall naturalness, \ding{193} ID consistency, \ding{194} text consistency, and \ding{195} audio-visual synchronization (ratings from minimum 1 to maximum 5). The results in \cref{tab:human} indicate that our method has a measurable advantage over these mainstream commercial models. 

\subsubsection{Application Scope Value}
We further evaluate {\method} in a variety of practical, complex application scenarios; as shown in \cref{fig:long}-Bottom, thanks to {\dataset} and our training strategies, {\method} can reliably generate long-term, semantically consistent animation videos, handling single-person and multi-person cases, singing, dancing, human interactions, large-scale motion (portrait, upper-body, and full-body), as well as anthropomorphic animations such as anime and animals.

\section{Conclusion} \label{sec:conclusion}
This paper presents {\method}, a framework for high-fidelity long-term digital human animation. It addresses challenges in datasets, long-term consistency, and efficient high-resolution generation. Leveraging the {\dataset} dataset and {\benchmark}, {\method} outperforms open-source and commercial models. With 11.4$\times$ inference speedup enabling real-time deployment, it helps bridge the gap between academia and digital human applications.

\noindent \textbf{Limitation and future work.} 
{\method} may produce artifacts when handling highly complex full-body motions, which remain a challenging problem in the field. We plan to expand the Soul-1M dataset to increase the representation of rare motion types and to include cross-lingual audio samples to improve model robustness. We will also investigate incorporating 3D geometric priors to enhance the naturalness and spatial consistency of full-body motions.




\section{Acknowledgements}
\noindent\textbf{Tencent Hunyuan.} 
Yi Chen, Qin Lin, Zeyi Lin, Qinglin Lu, Shuai Shao, Junshu Tang, Chunyu Wang, Hongmei Wang, Linqing Wang, Zhentao Yu, Tianxiang Zheng, Xin Zheng, Yuan Zhou, Zixiang Zhou

\noindent\textbf{Tencent AMS.} 
Haoyu Zhao, Junqi Cheng, Yuang Zhang, Jiaxi Gu, Zenghui Lu, Peng Shu

\noindent\textbf{Tencent Music Lyra Lab.} 
Yubin Zeng, Zhaokang Chen, Bin Wu


\setcitestyle{numbers,square}

\bibliography{youtu}

\begin{thebibliography}{93}
\providecommand{\natexlab}[1]{#1}
\providecommand{\url}[1]{\texttt{#1}}
\expandafter\ifx\csname urlstyle\endcsname\relax
  \providecommand{\doi}[1]{doi: #1}\else
  \providecommand{\doi}{doi: \begingroup \urlstyle{rm}\Url}\fi

\bibitem[Rekik et~al.(2024)Rekik, Wuhrer, Hoyet, Zibrek, and Olivier]{survey}
Rim Rekik, Stefanie Wuhrer, Ludovic Hoyet, Katja Zibrek, and Anne-H{\'e}l{\`e}ne Olivier.
\newblock A survey on realistic virtual human animations: Definitions, features and evaluations.
\newblock In \emph{Computer Graphics Forum}, 2024.

\bibitem[Gao et~al.(2025{\natexlab{a}})Gao, Hu, Hu, Huang, Ji, Meng, Qi, Qiao, Shen, Song, et~al.]{wans2v}
Xin Gao, Li~Hu, Siqi Hu, Mingyang Huang, Chaonan Ji, Dechao Meng, Jinwei Qi, Penchong Qiao, Zhen Shen, Yafei Song, et~al.
\newblock Wan-s2v: Audio-driven cinematic video generation.
\newblock \emph{arXiv preprint arXiv:2508.18621}, 2025{\natexlab{a}}.

\bibitem[Ding et~al.(2025)Ding, Liu, Zhang, Wang, Hu, Cui, Lao, Shao, Liu, Li, et~al.]{kling}
Yikang Ding, Jiwen Liu, Wenyuan Zhang, Zekun Wang, Wentao Hu, Liyuan Cui, Mingming Lao, Yingchao Shao, Hui Liu, Xiaohan Li, et~al.
\newblock Kling-avatar: Grounding multimodal instructions for cascaded long-duration avatar animation synthesis.
\newblock \emph{arXiv preprint arXiv:2509.09595}, 2025.

\bibitem[Gan et~al.(2025)Gan, Yang, Zhu, Xue, and Hoi]{omniavatar}
Qijun Gan, Ruizi Yang, Jianke Zhu, Shaofei Xue, and Steven Hoi.
\newblock Omniavatar: Efficient audio-driven avatar video generation with adaptive body animation.
\newblock \emph{arXiv preprint arXiv:2506.18866}, 2025.

\bibitem[Rombach et~al.(2022)Rombach, Blattmann, Lorenz, Esser, and Ommer]{ldm}
Robin Rombach, Andreas Blattmann, Dominik Lorenz, Patrick Esser, and Bj{\"o}rn Ommer.
\newblock High-resolution image synthesis with latent diffusion models.
\newblock In \emph{CVPR}, 2022.

\bibitem[Podell et~al.(2024)Podell, English, Lacey, Blattmann, Dockhorn, M{\"u}ller, Penna, and Rombach]{sdxl}
Dustin Podell, Zion English, Kyle Lacey, Andreas Blattmann, Tim Dockhorn, Jonas M{\"u}ller, Joe Penna, and Robin Rombach.
\newblock {SDXL}: Improving latent diffusion models for high-resolution image synthesis.
\newblock In \emph{ICLR}, 2024.

\bibitem[Esser et~al.(2024)Esser, Kulal, Blattmann, Entezari, M{\"u}ller, Saini, Levi, Lorenz, Sauer, Boesel, et~al.]{sd3}
Patrick Esser, Sumith Kulal, Andreas Blattmann, Rahim Entezari, Jonas M{\"u}ller, Harry Saini, Yam Levi, Dominik Lorenz, Axel Sauer, Frederic Boesel, et~al.
\newblock Scaling rectified flow transformers for high-resolution image synthesis.
\newblock In \emph{ICML}, 2024.

\bibitem[Labs(2024{\natexlab{a}})]{flux}
Black~Forest Labs.
\newblock Flux.
\newblock \url{https://github.com/black-forest-labs/flux}, 2024{\natexlab{a}}.

\bibitem[Labs et~al.(2025)Labs, Batifol, Blattmann, Boesel, Consul, Diagne, Dockhorn, English, English, Esser, Kulal, Lacey, Levi, Li, Lorenz, Müller, Podell, Rombach, Saini, Sauer, and Smith]{flux1kontext}
Black~Forest Labs, Stephen Batifol, Andreas Blattmann, Frederic Boesel, Saksham Consul, Cyril Diagne, Tim Dockhorn, Jack English, Zion English, Patrick Esser, Sumith Kulal, Kyle Lacey, Yam Levi, Cheng Li, Dominik Lorenz, Jonas Müller, Dustin Podell, Robin Rombach, Harry Saini, Axel Sauer, and Luke Smith.
\newblock Flux.1 kontext: Flow matching for in-context image generation and editing in latent space.
\newblock \emph{arXiv preprint arXiv:2506.15742}, 2025.

\bibitem[Guo et~al.(2024)Guo, Yang, Rao, Liang, Wang, Qiao, Agrawala, Lin, and Dai]{animatediff}
Yuwei Guo, Ceyuan Yang, Anyi Rao, Zhengyang Liang, Yaohui Wang, Yu~Qiao, Maneesh Agrawala, Dahua Lin, and Bo~Dai.
\newblock Animatediff: Animate your personalized text-to-image diffusion models without specific tuning.
\newblock In \emph{ICLR}, 2024.

\bibitem[Blattmann et~al.(2023)Blattmann, Dockhorn, Kulal, Mendelevitch, Kilian, Lorenz, Levi, English, Voleti, Letts, et~al.]{svd}
Andreas Blattmann, Tim Dockhorn, Sumith Kulal, Daniel Mendelevitch, Maciej Kilian, Dominik Lorenz, Yam Levi, Zion English, Vikram Voleti, Adam Letts, et~al.
\newblock Stable video diffusion: Scaling latent video diffusion models to large datasets.
\newblock \emph{arXiv preprint arXiv:2311.15127}, 2023.

\bibitem[Chen et~al.(2025{\natexlab{a}})Chen, Ge, Zhang, Zhang, Zhu, Yang, Hao, Wu, Lai, Hu, et~al.]{goku}
Shoufa Chen, Chongjian Ge, Yuqi Zhang, Yida Zhang, Fengda Zhu, Hao Yang, Hongxiang Hao, Hui Wu, Zhichao Lai, Yifei Hu, et~al.
\newblock Goku: Flow based video generative foundation models.
\newblock In \emph{CVPR}, 2025{\natexlab{a}}.

\bibitem[HaCohen et~al.(2024)HaCohen, Chiprut, Brazowski, Shalem, Moshe, Richardson, Levin, Shiran, Zabari, Gordon, Panet, Weissbuch, Kulikov, Bitterman, Melumian, and Bibi]{ltxvideo}
Yoav HaCohen, Nisan Chiprut, Benny Brazowski, Daniel Shalem, Dudu Moshe, Eitan Richardson, Eran Levin, Guy Shiran, Nir Zabari, Ori Gordon, Poriya Panet, Sapir Weissbuch, Victor Kulikov, Yaki Bitterman, Zeev Melumian, and Ofir Bibi.
\newblock Ltx-video: Realtime video latent diffusion.
\newblock \emph{arXiv preprint arXiv:2501.00103}, 2024.

\bibitem[Ma et~al.(2025{\natexlab{a}})Ma, Huang, Yan, Chen, Duan, Yin, Wan, Ming, Song, Chen, et~al.]{stepvideot2v}
Guoqing Ma, Haoyang Huang, Kun Yan, Liangyu Chen, Nan Duan, Shengming Yin, Changyi Wan, Ranchen Ming, Xiaoniu Song, Xing Chen, et~al.
\newblock Step-video-t2v technical report: The practice, challenges, and future of video foundation model.
\newblock \emph{arXiv preprint arXiv:2502.10248}, 2025{\natexlab{a}}.

\bibitem[Chen et~al.(2025{\natexlab{b}})Chen, Lin, Yang, Lin, Zhu, Fan, Zhang, Chen, Chen, Ma, et~al.]{skyreels}
Guibin Chen, Dixuan Lin, Jiangping Yang, Chunze Lin, Junchen Zhu, Mingyuan Fan, Hao Zhang, Sheng Chen, Zheng Chen, Chengcheng Ma, et~al.
\newblock Skyreels-v2: Infinite-length film generative model.
\newblock \emph{arXiv preprint arXiv:2504.13074}, 2025{\natexlab{b}}.

\bibitem[Kong et~al.(2024)Kong, Tian, Zhang, Min, Dai, Zhou, Xiong, Li, Wu, Zhang, et~al.]{hunyuanvideo}
Weijie Kong, Qi~Tian, Zijian Zhang, Rox Min, Zuozhuo Dai, Jin Zhou, Jiangfeng Xiong, Xin Li, Bo~Wu, Jianwei Zhang, et~al.
\newblock Hunyuanvideo: A systematic framework for large video generative models.
\newblock \emph{arXiv preprint arXiv:2412.03603}, 2024.

\bibitem[Wang et~al.(2025{\natexlab{a}})Wang, Ai, Wen, Mao, Xie, Chen, Yu, Zhao, Yang, Zeng, Wang, Zhang, Zhou, Wang, Chen, Zhu, Zhao, Yan, Huang, Feng, Zhang, Li, Wu, Chu, Feng, Zhang, Sun, Fang, Wang, Gui, Weng, Shen, Lin, Wang, Wang, Zhou, Wang, Shen, Yu, Shi, Huang, Xu, Kou, Lv, Li, Liu, Wang, Zhang, Huang, Li, Wu, Liu, Pan, Zheng, Hong, Shi, Feng, Jiang, Han, Wu, and Liu]{wan}
Ang Wang, Baole Ai, Bin Wen, Chaojie Mao, Chen-Wei Xie, Di~Chen, Feiwu Yu, Haiming Zhao, Jianxiao Yang, Jianyuan Zeng, Jiayu Wang, Jingfeng Zhang, Jingren Zhou, Jinkai Wang, Jixuan Chen, Kai Zhu, Kang Zhao, Keyu Yan, Lianghua Huang, Mengyang Feng, Ningyi Zhang, Pandeng Li, Pingyu Wu, Ruihang Chu, Ruili Feng, Shiwei Zhang, Siyang Sun, Tao Fang, Tianxing Wang, Tianyi Gui, Tingyu Weng, Tong Shen, Wei Lin, Wei Wang, Wei Wang, Wenmeng Zhou, Wente Wang, Wenting Shen, Wenyuan Yu, Xianzhong Shi, Xiaoming Huang, Xin Xu, Yan Kou, Yangyu Lv, Yifei Li, Yijing Liu, Yiming Wang, Yingya Zhang, Yitong Huang, Yong Li, You Wu, Yu~Liu, Yulin Pan, Yun Zheng, Yuntao Hong, Yupeng Shi, Yutong Feng, Zeyinzi Jiang, Zhen Han, Zhi-Fan Wu, and Ziyu Liu.
\newblock Wan: Open and advanced large-scale video generative models.
\newblock \emph{arXiv preprint arXiv:2503.20314}, 2025{\natexlab{a}}.

\bibitem[Xue et~al.(2025)Xue, Zhang, Hu, He, Chen, Cai, Wang, Wang, Liu, Li, et~al.]{ultravideo}
Zhucun Xue, Jiangning Zhang, Teng Hu, Haoyang He, Yinan Chen, Yuxuan Cai, Yabiao Wang, Chengjie Wang, Yong Liu, Xiangtai Li, et~al.
\newblock Ultravideo: High-quality uhd video dataset with comprehensive captions.
\newblock In \emph{NeurIPS}, 2025.

\bibitem[Hu et~al.(2026)Hu, Zhang, Su, and Yi]{ultragen}
Teng Hu, Jiangning Zhang, Zihan Su, and Ran Yi.
\newblock Ultragen: High-resolution video generation with hierarchical attention.
\newblock In \emph{AAAI}, 2026.

\bibitem[Bao et~al.(2024)Bao, Xiang, Yue, He, Zhu, Zheng, Zhao, Liu, Wang, and Zhu]{vidu}
Fan Bao, Chendong Xiang, Gang Yue, Guande He, Hongzhou Zhu, Kaiwen Zheng, Min Zhao, Shilong Liu, Yaole Wang, and Jun Zhu.
\newblock Vidu: a highly consistent, dynamic and skilled text-to-video generator with diffusion models.
\newblock \emph{arXiv preprint arXiv:2405.04233}, 2024.

\bibitem[Labs(2024{\natexlab{b}})]{ray2}
Luma Labs.
\newblock Ray2: Large-scale video generative model.
\newblock \url{https://lumalabs.ai/ray}, 2024{\natexlab{b}}.

\bibitem[Research(2023)]{runwaygen2}
Runway Research.
\newblock Gen-2: Generate novel videos with text, images or video clips.
\newblock Technical report, Runway ML, 02 2023.
\newblock URL \url{https://runwayml.com/research/gen-2}.

\bibitem[Limited(2025)]{pixversev5}
Moti Vai~Private Limited.
\newblock Pixverse v5.
\newblock \url{https://www.imagine.art/blogs/pixverse-v5-overview}, 2025.

\bibitem[Gao et~al.(2025{\natexlab{b}})Gao, Guo, Hoang, Huang, Jiang, Kong, Li, Li, Li, Li, et~al.]{seedance1}
Yu~Gao, Haoyuan Guo, Tuyen Hoang, Weilin Huang, Lu~Jiang, Fangyuan Kong, Huixia Li, Jiashi Li, Liang Li, Xiaojie Li, et~al.
\newblock Seedance 1.0: Exploring the boundaries of video generation models.
\newblock \emph{arXiv preprint arXiv:2506.09113}, 2025{\natexlab{b}}.

\bibitem[Zhang et~al.(2025{\natexlab{a}})Zhang, Yang, Zhang, Hu, Zhu, Lin, Mei, Jiang, Peng, and Yuan]{waver}
Yifu Zhang, Hao Yang, Yuqi Zhang, Yifei Hu, Fengda Zhu, Chuang Lin, Xiaofeng Mei, Yi~Jiang, Bingyue Peng, and Zehuan Yuan.
\newblock Waver: Wave your way to lifelike video generation.
\newblock \emph{arXiv preprint arXiv:2508.15761}, 2025{\natexlab{a}}.

\bibitem[Technology(2024)]{klingai}
Kuaishou Technology.
\newblock Kling ai - kuaishou's official ai video generation platform.
\newblock \url{https://klingai.com/cn/}, 2024.

\bibitem[Teng et~al.(2025)Teng, Jia, Sun, Li, Li, Tang, Han, Zhang, Zhang, Luo, et~al.]{magi}
Hansi Teng, Hongyu Jia, Lei Sun, Lingzhi Li, Maolin Li, Mingqiu Tang, Shuai Han, Tianning Zhang, WQ~Zhang, Weifeng Luo, et~al.
\newblock Magi-1: Autoregressive video generation at scale.
\newblock \emph{arXiv preprint arXiv:2505.13211}, 2025.

\bibitem[Wiedemer et~al.(2025)Wiedemer, Li, Vicol, Gu, Matarese, Swersky, Kim, Jaini, and Geirhos]{veo3}
Thadd{\"a}us Wiedemer, Yuxuan Li, Paul Vicol, Shixiang~Shane Gu, Nick Matarese, Kevin Swersky, Been Kim, Priyank Jaini, and Robert Geirhos.
\newblock Video models are zero-shot learners and reasoners.
\newblock \emph{arXiv preprint arXiv:2509.20328}, 2025.

\bibitem[Team(2025{\natexlab{a}})]{sora2}
OpenAI~Sora Team.
\newblock Sora 2.
\newblock \url{https://openai.com/zh-Hans-CN/index/sora-2/}, 2025{\natexlab{a}}.

\bibitem[Zhang et~al.(2023)Zhang, Cun, Wang, Zhang, Shen, Guo, Shan, and Wang]{sadtalker}
Wenxuan Zhang, Xiaodong Cun, Xuan Wang, Yong Zhang, Xi~Shen, Yu~Guo, Ying Shan, and Fei Wang.
\newblock Sadtalker: Learning realistic 3d motion coefficients for stylized audio-driven single image talking face animation.
\newblock In \emph{CVPR}, 2023.

\bibitem[Blanz and Vetter(2023)]{3dmm}
Volker Blanz and Thomas Vetter.
\newblock A morphable model for the synthesis of 3d faces.
\newblock In \emph{Seminal Graphics Papers: Pushing the Boundaries, Volume 2}, pages 157--164. 2023.

\bibitem[Goodfellow et~al.(2014)Goodfellow, Pouget-Abadie, Mirza, Xu, Warde-Farley, Ozair, Courville, and Bengio]{gan}
Ian~J. Goodfellow, Jean Pouget-Abadie, Mehdi Mirza, Bing Xu, David Warde-Farley, Sherjil Ozair, Aaron Courville, and Yoshua Bengio.
\newblock Generative adversarial nets.
\newblock In \emph{NeurIPS}, 2014.

\bibitem[Wei et~al.(2024)Wei, Yang, and Wang]{aniportrait}
Huawei Wei, Zejun Yang, and Zhisheng Wang.
\newblock Aniportrait: Audio-driven synthesis of photorealistic portrait animation.
\newblock \emph{arXiv preprint arXiv:2403.17694}, 2024.

\bibitem[Xu et~al.(2024)Xu, Li, Su, Shang, Zhang, Liu, Wang, Yao, and Zhu]{hallo}
Mingwang Xu, Hui Li, Qingkun Su, Hanlin Shang, Liwei Zhang, Ce~Liu, Jingdong Wang, Yao Yao, and Siyu Zhu.
\newblock Hallo: Hierarchical audio-driven visual synthesis for portrait image animation.
\newblock \emph{arXiv preprint arXiv:2406.08801}, 2024.

\bibitem[Cui et~al.(2024)Cui, Li, Yao, Zhu, Shang, Cheng, Zhou, Zhu, and Wang]{hallo2}
Jiahao Cui, Hui Li, Yao Yao, Hao Zhu, Hanlin Shang, Kaihui Cheng, Hang Zhou, Siyu Zhu, and Jingdong Wang.
\newblock Hallo2: Long-duration and high-resolution audio-driven portrait image animation.
\newblock \emph{arXiv preprint arXiv:2410.07718}, 2024.

\bibitem[Ji et~al.(2025)Ji, Hu, Xu, Zhu, Lin, He, Zhang, Luo, Chen, Lin, et~al.]{sonic}
Xiaozhong Ji, Xiaobin Hu, Zhihong Xu, Junwei Zhu, Chuming Lin, Qingdong He, Jiangning Zhang, Donghao Luo, Yi~Chen, Qin Lin, et~al.
\newblock Sonic: Shifting focus to global audio perception in portrait animation.
\newblock In \emph{CVPR}, 2025.

\bibitem[Yee et~al.(2025)Yee, Kollias, Mishra, and Dhall]{synchrorama}
Phyo~Thet Yee, Dimitrios Kollias, Sudeepta Mishra, and Abhinav Dhall.
\newblock Synchrorama: Lip-synchronized and emotion-aware talking face generation via multi-modal emotion embedding.
\newblock \emph{arXiv preprint arXiv:2509.19965}, 2025.

\bibitem[Nazarieh et~al.(2025)Nazarieh, Feng, Kanojia, Awais, and Kittler]{magictalk}
Fatemeh Nazarieh, Zhenhua Feng, Diptesh Kanojia, Muhammad Awais, and Josef Kittler.
\newblock Magic-talk: Motion-aware audio-driven talking face generation with customizable identity control.
\newblock \emph{arXiv preprint arXiv:2510.22810}, 2025.

\bibitem[Cui et~al.(2025)Cui, Li, Zhan, Shang, Cheng, Ma, Mu, Zhou, Wang, and Zhu]{hallo3}
Jiahao Cui, Hui Li, Yun Zhan, Hanlin Shang, Kaihui Cheng, Yuqi Ma, Shan Mu, Hang Zhou, Jingdong Wang, and Siyu Zhu.
\newblock Hallo3: Highly dynamic and realistic portrait image animation with video diffusion transformer.
\newblock In \emph{CVPR}, 2025.

\bibitem[Peebles and Xie(2023)]{dit}
William Peebles and Saining Xie.
\newblock Scalable diffusion models with transformers.
\newblock In \emph{ICCV}, 2023.

\bibitem[Liang et~al.(2025)Liang, Jiang, Liao, Yang, Zeng, Liang, et~al.]{alignhuman}
Chao Liang, Jianwen Jiang, Wang Liao, Jiaqi Yang, Weihong Zeng, Han Liang, et~al.
\newblock Alignhuman: Improving motion and fidelity via timestep-segment preference optimization for audio-driven human animation.
\newblock \emph{arXiv preprint arXiv:2506.11144}, 2025.

\bibitem[Seo et~al.(2025)Seo, Mira, Haliassos, Bounareli, Chen, Tran, Kim, Landgraf, and Shen]{lookaheadanchoring}
Junyoung Seo, Rodrigo Mira, Alexandros Haliassos, Stella Bounareli, Honglie Chen, Linh Tran, Seungryong Kim, Zoe Landgraf, and Jie Shen.
\newblock Lookahead anchoring: Preserving character identity in audio-driven human animation.
\newblock \emph{arXiv preprint arXiv:2510.23581}, 2025.

\bibitem[Zhang et~al.(2025{\natexlab{b}})Zhang, Wang, Jiang, Fan, Xu, and Qi]{fantasyid}
Yunpeng Zhang, Qiang Wang, Fan Jiang, Yaqi Fan, Mu~Xu, and Yonggang Qi.
\newblock Fantasyid: Face knowledge enhanced id-preserving video generation.
\newblock \emph{arXiv preprint arXiv:2502.13995}, 2025{\natexlab{b}}.

\bibitem[Wang et~al.(2025{\natexlab{b}})Wang, Wang, Jiang, Fan, Zhang, Qi, Zhao, and Xu]{fantasytalking}
Mengchao Wang, Qiang Wang, Fan Jiang, Yaqi Fan, Yunpeng Zhang, Yonggang Qi, Kun Zhao, and Mu~Xu.
\newblock Fantasytalking: Realistic talking portrait generation via coherent motion synthesis.
\newblock \emph{arXiv preprint arXiv:2504.04842}, 2025{\natexlab{b}}.

\bibitem[Chen et~al.(2025{\natexlab{c}})Chen, Liang, Zhou, Huang, Ma, Tang, Lin, Zhou, and Lu]{hunyuanvideoavatar}
Yi~Chen, Sen Liang, Zixiang Zhou, Ziyao Huang, Yifeng Ma, Junshu Tang, Qin Lin, Yuan Zhou, and Qinglin Lu.
\newblock Hunyuanvideo-avatar: High-fidelity audio-driven human animation for multiple characters.
\newblock \emph{arXiv preprint arXiv:2505.20156}, 2025{\natexlab{c}}.

\bibitem[Li et~al.(2025)Li, Xie, Ren, Gan, Zhang, Kong, Yin, Peng, and Yuan]{infinityhuman}
Xiaodi Li, Pan Xie, Yi~Ren, Qijun Gan, Chen Zhang, Fangyuan Kong, Xiang Yin, Bingyue Peng, and Zehuan Yuan.
\newblock Infinityhuman: Towards long-term audio-driven human.
\newblock \emph{arXiv preprint arXiv:2508.20210}, 2025.

\bibitem[Yang et~al.(2025{\natexlab{a}})Yang, Teng, Zheng, Ding, Huang, Xu, Yang, Hong, Zhang, Feng, Yin, Yuxuan.Zhang, Wang, Cheng, Xu, Gu, Dong, and Tang]{cogvideox}
Zhuoyi Yang, Jiayan Teng, Wendi Zheng, Ming Ding, Shiyu Huang, Jiazheng Xu, Yuanming Yang, Wenyi Hong, Xiaohan Zhang, Guanyu Feng, Da~Yin, Yuxuan.Zhang, Weihan Wang, Yean Cheng, Bin Xu, Xiaotao Gu, Yuxiao Dong, and Jie Tang.
\newblock Cogvideox: Text-to-video diffusion models with an expert transformer.
\newblock In \emph{ICLR}, 2025{\natexlab{a}}.

\bibitem[Meng et~al.(2025{\natexlab{a}})Meng, Wang, Wu, Zheng, Li, and Ma]{echomimicv3}
Rang Meng, Yan Wang, Weipeng Wu, Ruobing Zheng, Yuming Li, and Chenguang Ma.
\newblock Echomimicv3: 1.3 b parameters are all you need for unified multi-modal and multi-task human animation.
\newblock \emph{arXiv preprint arXiv:2507.03905}, 2025{\natexlab{a}}.

\bibitem[Tu et~al.(2025)Tu, Pan, Huang, Han, Xing, Dai, Luo, Wu, and Jiang]{stableavatar}
Shuyuan Tu, Yueming Pan, Yinming Huang, Xintong Han, Zhen Xing, Qi~Dai, Chong Luo, Zuxuan Wu, and Yu-Gang Jiang.
\newblock Stableavatar: Infinite-length audio-driven avatar video generation.
\newblock \emph{arXiv preprint arXiv:2508.08248}, 2025.

\bibitem[Hu et~al.(2025)Hu, Yu, Zhou, Liang, Zhou, Lin, and Lu]{hunyuancustom}
Teng Hu, Zhentao Yu, Zhengguang Zhou, Sen Liang, Yuan Zhou, Qin Lin, and Qinglin Lu.
\newblock Hunyuancustom: A multimodal-driven architecture for customized video generation.
\newblock \emph{arXiv preprint arXiv:2505.04512}, 2025.

\bibitem[Jiang et~al.(2025)Jiang, Zeng, Zheng, Yang, Liang, Liao, Liang, Zhang, and Gao]{omnihuman}
Jianwen Jiang, Weihong Zeng, Zerong Zheng, Jiaqi Yang, Chao Liang, Wang Liao, Han Liang, Yuan Zhang, and Mingyuan Gao.
\newblock Omnihuman-1.5: Instilling an active mind in avatars via cognitive simulation.
\newblock \emph{arXiv preprint arXiv:2508.19209}, 2025.

\bibitem[Wang et~al.(2024)Wang, Tian, Zhang, Guan, Luo, Shen, Jiang, Gu, Han, and Yang]{vexpress}
Cong Wang, Kuan Tian, Jun Zhang, Yonghang Guan, Feng Luo, Fei Shen, Zhiwei Jiang, Qing Gu, Xiao Han, and Wei Yang.
\newblock V-express: Conditional dropout for progressive training of portrait video generation.
\newblock \emph{arXiv preprint arXiv:2406.02511}, 2024.

\bibitem[Chen et~al.(2025{\natexlab{d}})Chen, Cao, Chen, Li, and Ma]{echomimic}
Zhiyuan Chen, Jiajiong Cao, Zhiquan Chen, Yuming Li, and Chenguang Ma.
\newblock Echomimic: Lifelike audio-driven portrait animations through editable landmark conditions.
\newblock In \emph{AAAI}, 2025{\natexlab{d}}.

\bibitem[Meng et~al.(2025{\natexlab{b}})Meng, Zhang, Li, and Ma]{echomimicv2}
Rang Meng, Xingyu Zhang, Yuming Li, and Chenguang Ma.
\newblock Echomimicv2: Towards striking, simplified, and semi-body human animation.
\newblock In \emph{CVPR}, 2025{\natexlab{b}}.

\bibitem[Chen et~al.(2025{\natexlab{e}})Chen, Cui, Zhang, Zhang, Zhou, Li, Tang, Liu, Liao, Chen, et~al.]{midas}
Ming Chen, Liyuan Cui, Wenyuan Zhang, Haoxian Zhang, Yan Zhou, Xiaohan Li, Songlin Tang, Jiwen Liu, Borui Liao, Hejia Chen, et~al.
\newblock Midas: Multimodal interactive digital-human synthesis via real-time autoregressive video generation.
\newblock \emph{arXiv preprint arXiv:2508.19320}, 2025{\natexlab{e}}.

\bibitem[Huang et~al.(2025)Huang, Li, Liu, Huang, Yang, Wang, and Xu]{vividanimator}
Donglin Huang, Yongyuan Li, Tianhang Liu, Junming Huang, Xiaoda Yang, Chi Wang, and Weiwei Xu.
\newblock Vividanimator: An end-to-end audio and pose-driven half-body human animation framework.
\newblock \emph{arXiv preprint arXiv:2510.10269}, 2025.

\bibitem[Zhu et~al.(2025)Zhu, Yu, Wang, Sun, and Zheng]{egstalker}
Tianheng Zhu, Yinfeng Yu, Liejun Wang, Fuchun Sun, and Wendong Zheng.
\newblock Egstalker: Real-time audio-driven talking head generation with efficient gaussian deformation.
\newblock \emph{arXiv preprint arXiv:2510.08587}, 2025.

\bibitem[Su et~al.(2025)Su, Wei, Li, Yang, and Deng]{muex}
Zibo Su, Kun Wei, Jiahua Li, Xu~Yang, and Cheng Deng.
\newblock A bridge from audio to video: Phoneme-viseme alignment allows every face to speak multiple languages.
\newblock \emph{arXiv preprint arXiv:2510.06612}, 2025.

\bibitem[Deng et~al.(2025)Deng, Wu, Zheng, Zhang, He, and Han]{avatarsync}
Yuchen Deng, Xiuyang Wu, Hai-Tao Zheng, Suiyang Zhang, Yi~He, and Yuxing Han.
\newblock Avatarsync: Rethinking talking-head animation through autoregressive perspective.
\newblock \emph{arXiv preprint arXiv:2509.12052}, 2025.

\bibitem[Wang et~al.(2025{\natexlab{c}})Wang, Wang, Jiang, Fan, Qi, and Xu]{fantasyportrait}
Qiang Wang, Mengchao Wang, Fan Jiang, Yaqi Fan, Yonggang Qi, and Mu~Xu.
\newblock Fantasyportrait: Enhancing multi-character portrait animation with expression-augmented diffusion transformers.
\newblock \emph{arXiv preprint arXiv:2507.12956}, 2025{\natexlab{c}}.

\bibitem[Kong et~al.(2025)Kong, Gao, Zhang, Kang, Wei, Cai, Chen, and Luo]{multitalk}
Zhe Kong, Feng Gao, Yong Zhang, Zhuoliang Kang, Xiaoming Wei, Xunliang Cai, Guanying Chen, and Wenhan Luo.
\newblock Let them talk: Audio-driven multi-person conversational video generation.
\newblock \emph{arXiv preprint arXiv:2505.22647}, 2025.

\bibitem[Ma et~al.(2025{\natexlab{b}})Ma, Huang, Cai, Guan, Zheng, Zhao, Zhang, and Zhang]{playmate2}
Xingpei Ma, Shenneng Huang, Jiaran Cai, Yuansheng Guan, Shen Zheng, Hanfeng Zhao, Qiang Zhang, and Shunsi Zhang.
\newblock Playmate2: Training-free multi-character audio-driven animation via diffusion transformer with reward feedback.
\newblock \emph{arXiv preprint arXiv:2510.12089}, 2025{\natexlab{b}}.

\bibitem[Dao et~al.(2022)Dao, Fu, Ermon, Rudra, and R{\'e}]{flashattention1}
Tri Dao, Dan Fu, Stefano Ermon, Atri Rudra, and Christopher R{\'e}.
\newblock Flashattention: Fast and memory-efficient exact attention with io-awareness.
\newblock \emph{NeurIPS}, 2022.

\bibitem[Dao(2023)]{flashattention2}
Tri Dao.
\newblock Flashattention-2: Faster attention with better parallelism and work partitioning.
\newblock \emph{arXiv preprint arXiv:2307.08691}, 2023.

\bibitem[Shah et~al.(2024)Shah, Bikshandi, Zhang, Thakkar, Ramani, and Dao]{flashattention3}
Jay Shah, Ganesh Bikshandi, Ying Zhang, Vijay Thakkar, Pradeep Ramani, and Tri Dao.
\newblock Flashattention-3: Fast and accurate attention with asynchrony and low-precision.
\newblock In \emph{NeurIPS}, 2024.

\bibitem[Zhang et~al.(2025{\natexlab{c}})Zhang, Wei, Zhang, Zhu, and Chen]{sageattention1}
Jintao Zhang, Jia Wei, Pengle Zhang, Jun Zhu, and Jianfei Chen.
\newblock Sageattention: Accurate 8-bit attention for plug-and-play inference acceleration.
\newblock In \emph{ICLR}, 2025{\natexlab{c}}.

\bibitem[Zhang et~al.(2025{\natexlab{d}})Zhang, Huang, Zhang, Wei, Zhu, and Chen]{sageattention2}
Jintao Zhang, Haofeng Huang, Pengle Zhang, Jia Wei, Jun Zhu, and Jianfei Chen.
\newblock Sageattention2: Efficient attention with thorough outlier smoothing and per-thread int4 quantization.
\newblock In \emph{ICML}, 2025{\natexlab{d}}.

\bibitem[Zhang et~al.(2025{\natexlab{e}})Zhang, Wei, Zhang, Xu, Huang, Wang, Jiang, Zhu, and Chen]{sageattention3}
Jintao Zhang, Jia Wei, Pengle Zhang, Xiaoming Xu, Haofeng Huang, Haoxu Wang, Kai Jiang, Jun Zhu, and Jianfei Chen.
\newblock Sageattention3: Microscaling fp4 attention for inference and an exploration of 8-bit training.
\newblock In \emph{NeurIPS}, 2025{\natexlab{e}}.

\bibitem[Zhang et~al.(2025{\natexlab{f}})Zhang, Xiang, Huang, Xi, Zhu, Chen, et~al.]{spargeattention}
Jintao Zhang, Chendong Xiang, Haofeng Huang, Haocheng Xi, Jun Zhu, Jianfei Chen, et~al.
\newblock Spargeattention: Accurate and training-free sparse attention accelerating any model inference.
\newblock In \emph{ICML}, 2025{\natexlab{f}}.

\bibitem[Zhang et~al.(2025{\natexlab{g}})Zhang, Chen, Huang, Lin, Liu, Stoica, Xing, and Zhang]{vsa}
Peiyuan Zhang, Yongqi Chen, Haofeng Huang, Will Lin, Zhengzhong Liu, Ion Stoica, Eric Xing, and Hao Zhang.
\newblock Vsa: Faster video diffusion with trainable sparse attention.
\newblock \emph{arXiv preprint arXiv:2505.13389}, 2025{\natexlab{g}}.

\bibitem[Liu et~al.(2025)Liu, Zhang, Li, Bai, Han, Tang, Xing, Wu, Yang, Chen, et~al.]{fpsattention}
Akide Liu, Zeyu Zhang, Zhexin Li, Xuehai Bai, Yizeng Han, Jiasheng Tang, Yuanjie Xing, Jichao Wu, Mingyang Yang, Weihua Chen, et~al.
\newblock Fpsattention: Training-aware fp8 and sparsity co-design for fast video diffusion.
\newblock \emph{arXiv preprint arXiv:2506.04648}, 2025.

\bibitem[Radford et~al.(2023)Radford, Kim, Xu, Brockman, McLeavey, and Sutskever]{whisper}
Alec Radford, Jong~Wook Kim, Tao Xu, Greg Brockman, Christine McLeavey, and Ilya Sutskever.
\newblock Robust speech recognition via large-scale weak supervision.
\newblock In \emph{ICML}, 2023.

\bibitem[Yin et~al.(2024)Yin, Gharbi, Park, Zhang, Shechtman, Durand, and Freeman]{DMD2}
Tianwei Yin, Micha{\"e}l Gharbi, Taesung Park, Richard Zhang, Eli Shechtman, Fredo Durand, and Bill Freeman.
\newblock Improved distribution matching distillation for fast image synthesis.
\newblock \emph{NeurIPS}, 2024.

\bibitem[Chung et~al.(2018)Chung, Nagrani, and Zisserman]{voxceleb2}
Joon~Son Chung, Arsha Nagrani, and Andrew Zisserman.
\newblock Voxceleb2: Deep speaker recognition.
\newblock \emph{arXiv preprint arXiv:1806.05622}, 2018.

\bibitem[Xie et~al.(2022)Xie, Wang, Zhang, Dong, and Shan]{vfhq}
Liangbin Xie, Xintao Wang, Honglun Zhang, Chao Dong, and Ying Shan.
\newblock Vfhq: A high-quality dataset and benchmark for video face super-resolution.
\newblock In \emph{CVPR}, 2022.

\bibitem[Castellano(2024)]{pyscenedetect}
Brandon Castellano.
\newblock Pyscenedetect: Python-based video scene detector, March 2024.
\newblock URL \url{https://github.com/Breakthrough/PySceneDetect}.
\newblock Accessed: [Insert Date].

\bibitem[Oquab et~al.(2023)Oquab, Darcet, Moutakanni, Vo, Szafraniec, Khalidov, Fernandez, Haziza, Massa, El-Nouby, et~al.]{dinov2}
Maxime Oquab, Timoth{\'e}e Darcet, Th{\'e}o Moutakanni, Huy Vo, Marc Szafraniec, Vasil Khalidov, Pierre Fernandez, Daniel Haziza, Francisco Massa, Alaaeldin El-Nouby, et~al.
\newblock Dinov2: Learning robust visual features without supervision.
\newblock \emph{arXiv preprint arXiv:2304.07193}, 2023.

\bibitem[Deng et~al.(2020)Deng, Guo, Ververas, Kotsia, and Zafeiriou]{retinaface}
Jiankang Deng, Jia Guo, Evangelos Ververas, Irene Kotsia, and Stefanos Zafeiriou.
\newblock Retinaface: Single-shot multi-level face localisation in the wild.
\newblock In \emph{CVPR}, 2020.

\bibitem[Duan et~al.(2025)Duan, Hu, Wang, Yang, Xu, Liu, Min, Cai, Ye, Zhang, et~al.]{finevq}
Huiyu Duan, Qiang Hu, Jiarui Wang, Liu Yang, Zitong Xu, Lu~Liu, Xiongkuo Min, Chunlei Cai, Tianxiao Ye, Xiaoyun Zhang, et~al.
\newblock Finevq: Fine-grained user generated content video quality assessment.
\newblock In \emph{CVPR}, 2025.

\bibitem[PaddlePaddle(2023)]{paddleocr}
PaddlePaddle.
\newblock Paddleocr.
\newblock \url{https://github.com/PaddlePaddle/PaddleOCR}, 2023.

\bibitem[Bai et~al.(2025)Bai, Chen, Liu, Wang, Ge, Song, Dang, Wang, Wang, Tang, et~al.]{qwen25vl}
Shuai Bai, Keqin Chen, Xuejing Liu, Jialin Wang, Wenbin Ge, Sibo Song, Kai Dang, Peng Wang, Shijie Wang, Jun Tang, et~al.
\newblock Qwen2.5-vl technical report.
\newblock \emph{arXiv preprint arXiv:2502.13923}, 2025.

\bibitem[Contributors(2020)]{mmpose}
MMPose Contributors.
\newblock Openmmlab pose estimation toolbox and benchmark.
\newblock \url{https://github.com/open-mmlab/mmpose}, 2020.

\bibitem[Chung and Zisserman(2016)]{syncnet}
Joon~Son Chung and Andrew Zisserman.
\newblock Out of time: automated lip sync in the wild.
\newblock In \emph{ACCV}, 2016.

\bibitem[Team(2025{\natexlab{b}})]{qwen3}
Qwen Team.
\newblock Qwen3, April 2025{\natexlab{b}}.
\newblock URL \url{https://qwenlm.github.io/blog/qwen3/}.

\bibitem[Yariv et~al.(2024)Yariv, Gat, Benaim, Wolf, Schwartz, and Adi]{av_align}
Guy Yariv, Itai Gat, Sagie Benaim, Lior Wolf, Idan Schwartz, and Yossi Adi.
\newblock Diverse and aligned audio-to-video generation via text-to-video model adaptation.
\newblock In \emph{AAAI}, 2024.

\bibitem[Yang et~al.(2025{\natexlab{b}})Yang, Kong, Gao, Cheng, Liu, Zhang, Kang, Luo, Cai, He, et~al.]{infinitetalk}
Shaoshu Yang, Zhe Kong, Feng Gao, Meng Cheng, Xiangyu Liu, Yong Zhang, Zhuoliang Kang, Wenhan Luo, Xunliang Cai, Ran He, et~al.
\newblock Infinitetalk: Audio-driven video generation for sparse-frame video dubbing.
\newblock \emph{arXiv preprint arXiv:2508.14033}, 2025{\natexlab{b}}.

\bibitem[Cao et~al.(2025)Cao, Chen, Chen, Cheng, Cui, Deng, Dong, Gong, Gu, Gu, et~al.]{hunyuanimage3}
Siyu Cao, Hangting Chen, Peng Chen, Yiji Cheng, Yutao Cui, Xinchi Deng, Ying Dong, Kipper Gong, Tianpeng Gu, Xiusen Gu, et~al.
\newblock Hunyuanimage 3.0 technical report.
\newblock \emph{arXiv preprint arXiv:2509.23951}, 2025.

\bibitem[Zhou et~al.(2025)Zhou, Zhou, He, Zhou, Wang, Deng, and Shu]{indextts2}
Siyi Zhou, Yiquan Zhou, Yi~He, Xun Zhou, Jinchao Wang, Wei Deng, and Jingchen Shu.
\newblock Indextts2: A breakthrough in emotionally expressive and duration-controlled auto-regressive zero-shot text-to-speech.
\newblock \emph{arXiv preprint arXiv:2506.21619}, 2025.

\bibitem[Guo et~al.(2025)Guo, Yang, Zhang, Song, Zhang, Xu, Zhu, Ma, Wang, Bi, et~al.]{deepseekr1}
Daya Guo, Dejian Yang, Haowei Zhang, Junxiao Song, Ruoyu Zhang, Runxin Xu, Qihao Zhu, Shirong Ma, Peiyi Wang, Xiao Bi, et~al.
\newblock Deepseek-r1: Incentivizing reasoning capability in llms via reinforcement learning.
\newblock \emph{arXiv preprint arXiv:2501.12948}, 2025.

\bibitem[Deng et~al.(2019)Deng, Guo, Niannan, and Zafeiriou]{arcface}
Jiankang Deng, Jia Guo, Xue Niannan, and Stefanos Zafeiriou.
\newblock Arcface: Additive angular margin loss for deep face recognition.
\newblock In \emph{CVPR}, 2019.

\bibitem[hey()]{heygen}
Heygen home page.
\newblock \url{https://app.heygen.com/home}.
\newblock Accessed: 2025-11-1].

\bibitem[Loshchilov and Hutter(2019)]{adamw}
Ilya Loshchilov and Frank Hutter.
\newblock Decoupled weight decay regularization.
\newblock In \emph{ICLR}, 2019.

\bibitem[Li et~al.(2024)Li, Zhang, Xu, Lin, Xie, Feng, Peng, Chen, and Xing]{latentsync}
Chunyu Li, Chao Zhang, Weikai Xu, Jingyu Lin, Jinghui Xie, Weiguo Feng, Bingyue Peng, Cunjian Chen, and Weiwei Xing.
\newblock Latentsync: Taming audio-conditioned latent diffusion models for lip sync with syncnet supervision.
\newblock \emph{arXiv preprint arXiv:2412.09262}, 2024.

\end{thebibliography}

\end{document}